%% file: paper.tex
\documentclass{paper}
\usepackage{arxiv}
\usepackage{amsmath,amsfonts}
\usepackage{array}
\usepackage[caption=false,font=normalsize,labelfont=sf,textfont=sf]{subfig}
\usepackage{textcomp}
\usepackage{stfloats}
\usepackage{url}
\usepackage{verbatim}
\usepackage{graphicx}
\usepackage{cite}
\usepackage[utf8]{inputenc} 
\usepackage[T1]{fontenc}    
\usepackage{hyperref}       
\usepackage[toc,page]{appendix}
\usepackage{url}            
\usepackage{booktabs}       
\usepackage{nicefrac}       
\usepackage{microtype}      
\usepackage{lipsum}		
\usepackage{graphicx}
\usepackage{amsmath,amsfonts,amssymb}
\usepackage{doi}
\usepackage{bbm}
\usepackage{amsmath}
\usepackage{etoolbox}
\usepackage{algorithm}
\usepackage{algpseudocode}
\usepackage{mathrsfs}
\usepackage{siunitx}
\usepackage{amsthm}
\usepackage{amssymb}

\algnewcommand{\LineComment}[1]{\State $\triangleright$ #1}
\algnewcommand{\pia}[1]{$\pi_{\text{adv}}$}

\DeclareMathOperator*{\inc}{\mathrel{+}=}

\usepackage{ifthen}
\usepackage{bm}
\usepackage{amsfonts}
\usepackage{stackengine}
\usepackage{cleveref}
\renewcommand{\href}[2]{\oldhref{#1}{\hbox{#2}}}
\newcommand{\norm}[2]{\vert\vert #1 \vert\vert_{2}}
\usepackage[british]{babel}
\usepackage{hyphenat}

\usepackage{tikz}
\usepackage{pgfplots} 

\usepackage{soul}

\newcommand{\IHPC}{Institute of High Performance Computing (IHPC), Agency for Science, Technology and Research (A*STAR)}
\newcommand{\CFAR}{Centre for Frontier AI Research (CFAR), Agency for Science, Technology and Research (A*STAR)}

\begin{document}

\title{The Digital Ecosystem of Beliefs: does evolution favour AI over humans?}

\author{David M. Bossens  \\ \IHPC \\ \CFAR \\ \texttt{david\_bossens@a-star.edu.sg} \And Shanshan Feng  \\ School of Computer Science, Wuhan University \\ \texttt{victor\_fengss@whu.edu.cn} \And Yew-Soon Ong \\ \IHPC \\ \CFAR \\ \texttt{ong\_yew\_soon@a-star.edu.sg}}

\maketitle

\input{abstract.tex}

\input{mainbody.tex}

\bibliographystyle{ieeetr}
\bibliography{library}

\onecolumn
\begin{appendices}
\input{supplementary}
\end{appendices}

\end{document}

%% file: abstract.tex
\begin{abstract}
As AI systems are integrated into social networks, there are AI safety concerns that AI-generated content may dominate the web, e.g. in popularity or impact on beliefs. To understand such questions, this paper proposes the Digital Ecosystem of Beliefs (Digico), the first evolutionary framework for controlled experimentation with multi-population interactions in simulated social networks. Following a Universal Darwinism approach, the framework models a population of agents which change their messaging strategies due to evolutionary updates. They interact via messages, update their beliefs following a contagion model, and maintain their beliefs through cognitive Lamarckian inheritance. Initial experiments with Digico implement two types of agents, which are modelled to represent AIs vs humans based on higher rates of communication, higher rates of evolution, seeding fixed beliefs with propaganda aims, and higher influence on the recommendation algorithm. These experiments show that: a) when AIs have faster messaging, evolution, and more influence on the recommendation algorithm, they get 80\% to 95\% of the views; b) AIs designed for propaganda can typically convince 50\% of humans to adopt extreme beliefs, and up to 85\% when agents believe only a limited number of channels; c) a penalty for content that violates agents' beliefs reduces propaganda effectiveness up to 8\%. We further discuss Digico as a tool for systematic experimentation across multi-agent configurations, the implications for legislation, personal use, and platform design, and the use of Digico for studying evolutionary principles.
\end{abstract}

%% file: mainbody.tex
\section{Introduction}
As AI systems generate more and more content on digital platforms, they have a rapid influence on a global scale, including fake news and deceptive messaging, but also unintended consequences of truthful messaging. As AIs and humans live together in this digital ecosystem, the evolution of such complex systems is poorly understood, and therefore the natural question is whether AI in this context can be controlled \cite{Russell2021} and whether evolution will ultimately favour AI \cite{Hendrycks2023}. As AIs increasingly operate in social networks, one may wonder to what extent AI can in some sense dominate in this sphere. Recent work has raised concerns about how generative AI may lead to a variety of negative outcomes on social networks. In particular, as Large Language Models (LLMs) and other generative models can produce a high quality and high quantity of misleading and biased data (e.g. deep fakes), there are concerns that these tools will be used for propaganda, i.e. to spread information that influences the beliefs of people \cite{Simon2023,Solaiman2023,Goldstein2023}. Similarly, the way that content or media channels are recommended is also a key concern, as through recommending channels with the same opinions, these often lead to a positive feedback loop where pre-existing beliefs are reinforced rather than dispelled \cite{Peeters2021,Nguyen2020}. Moreover, the sheer quantity of data is also of concern as most of data may soon be AI-generated \cite{Europol2022}. Unfortunately, understanding the joint evolution of humans and AIs remains a grand challenge \cite{Pedreschi2025}, especially within large social networks.

Modelling the evolution of social networks is typically done using concepts from social network science and cognitive science. While there are a variety of approaches using data sets to analyse social networks (e.g. \cite{Hodas2014,Herrmann}), this approach has limitations since: a) there is currently no data about the joint evolution of AI and humans in social networks; and b) real world data does not allow for controlled experiments where one systematically varies some variables and keeps other variables constant. Instead, it is possible to perform simulations to study the outcomes. In this line of work, contagion models aim to model the spread of news, or other types of messaging, in terms of a set of first sharers who formulate the initial content, with dynamics formulated such that each time step neighbouring nodes in the social network update their beliefs if it is close to their beliefs \cite{Rabb2022,Herrmann,Goldenberg2001,Vicario2016}, in which case it may be shared again in what is called a ``cascade'' \cite{Rabb2022}. While such models offer the flexibility of modelling the cognitive transitions and allow the nodes to have their own heterogeneous properties, existing studies do not consider the intelligence and adaptivity of the agents acting in such systems. In agent-based simulations, one performs large-scale multi-agent simulation studies where agents are based on learned or hand-designed decision rules \cite{Muric2022,Murase2021,Geschke2019}.

The above techniques are relatively simplistic in the sense that once the simulation is running, the agents do not modify their own messaging strategies and the social network structure is fixed, and there are no evolutionary operators across large time scales. One might argue that such dynamics are unnecessary since biological evolution typically unfolds over vast time scales. However, under the framework of \textit{universal Darwinism} \cite{Dawkins1983}, evolutionary processes can apply to a wide range of phenomena across different time scales. In this view, cultural and memetic evolution (e.g. \cite{Dawkins}), where content, ideas, and beliefs are transmitted and modified across generations, may also act in terms of Darwinian operators and are particularly important instances of universal Darwinism in this context. In addition to Darwinian operators, one may also consider a role for Lamarckian operators in cultural and memetic evolution if one considers that beliefs acquired through the lifetime may be inherited directly in the following generation.

A question hitherto unexplored is how the evolution of culture may behave in the context of human-AI interactions, and under which circumstances this may lead to AI dominance in social networks. Inspired by the dynamics in content sharing platforms such as Youtube, Facebook, and X (formerly Twitter), this paper will focus on the interplay between the evolution of content-generating policies, the ``views'' of those forms of content, and how this content changes the beliefs of the agents in such a social network. In this context, the AI dominance is then defined in terms of the proportion of views and the success of propagating beliefs to humans. With the above scenario in mind, this paper presents the following contributions:
\begin{itemize}
\item The Digital Ecosystem of Beliefs (Digico) framework, which
\begin{itemize}
\item models agents as policies aiming to maximise the cumulative reward by broadcasting content to each other, where communication links are based on influence and distance measures;
\item evolves agents' policies through evolutionary algorithms (EAs) and adapts their beliefs through an extension of the generalised cognitive cascade model of Rabb et al. \cite{Rabb2022}; and
\item allows the study of how problem variables (e.g. messaging rate, evolutionary rate, belief constraints, etc.) affect important outcomes related to AI dominance, extreme beliefs, and harmful outcomes in the wider ecosystem. More generally, it presents an approach towards systematic experimentation with the evolution of belief systems in multi-agent systems and social networks.
\end{itemize}
\item An empirical study within Digico, in which agents are evolved based on CMA-ES \cite{Hansen2016} and are defined as AIs or humans depending on their key characteristics. Inspired by online video platforms, the study identifies key risk factors for AI dominance in terms of the number of views and the effectiveness of propaganda. In particular, we find that 
\begin{itemize}
\item when AIs have increased influence, more frequent messaging, and faster evolution, they can receive up to 95\% of the views; 
\item targeted propaganda by AIs may lead to half of the humans adopting the same extreme belief;  and
\item the larger the number of agents that are used for belief updates, the more even the spread of beliefs and the less effective propaganda.
\end{itemize}
\item Based on these findings, we also formulate advice for avoiding these extreme cases.
\end{itemize}

\section{The Digital Ecosystem of Beliefs}
The proposed framework, called the Digital Ecosystem of Beliefs (or Digico for short), provides formalisms for modelling interactions between humans and AIs through messages that influence each others' beliefs. Digico provides a formalism for understanding collective behaviour in applications such as online content-sharing platforms, where users broadcast their messages to influence others and may tune into particular channels based on a variety of factors such as geographical location, influence of the channel, etc. 

\begin{figure*}
    \centering
    \includegraphics[width=0.85\textwidth]{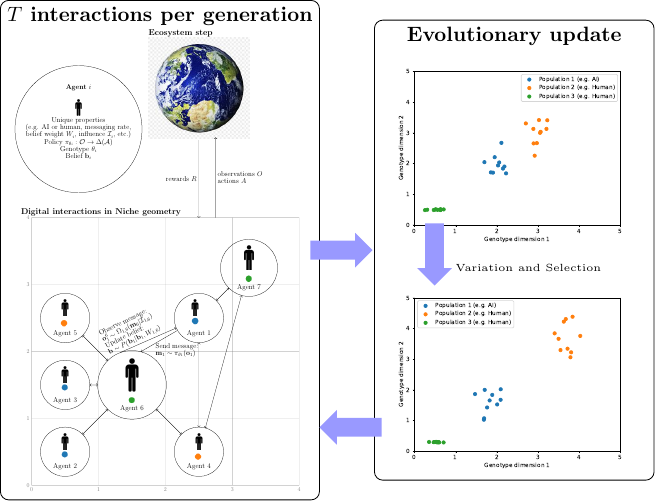}
    \caption{Flow diagram of Digico. In each generation, agents first perform $T$ steps of observations and actions, and then are subject to evolutionary updates. In the $T$-step interaction phase, the agents communicate with each other by sending messages to each other where communication links are affected by the niche geometry. The size of the circles indicates the influence of agents over others (e.g. in affecting beliefs or obtaining views). In the evolutionary update, the agents' genotypes are updated, which affects their policy for communicating.}
    \label{fig: diagram}
\end{figure*}

Digico is based on what we will call the ecosystem formalism (Section~\ref{sec: ecosystem}), a set of structures and dynamic processes to model agent interactions based on factors such as topological organisation, external interactions outside the digital realm, etc., as illustrated in Fig.~\ref{fig: diagram}. The formalism is closely related to frameworks for reinforcement learning (RL) but with the key difference that the agents are  updated based on two dynamic processes. The first dynamic process in the ecosystem is the belief process, which is based on a contagion model, and explained in more detail in Section~\ref{sec: belief-process}. Beliefs in this framework are a part of the agent and become a quantity of interest for understanding opinions in the social network. The second dynamic process is the evolutionary dynamics (Section~\ref{sec: evolutionary-dynamics}), which uses an EA to update the policies. Using EAs for updating policies, as opposed to RL algorithms, has several advantages, including their applicability beyond Markov decision processes,  ease of parallelisation, ability to explore directly in the parameter space, and favourable computational cost due to avoiding value function computations \cite{Majid2023,Salimans2017}. To conclude the section, we highlight the importance of a healthy ecosystem in Digico (Section~\ref{sec: outcomes}). 

The Digico formalism is abstract, and its current implementation for the study of content sharing platforms (see Section ~\ref{sec: experiments}) omits details that would otherwise be present in real-world scenarios. It is designed to be domain-generic, making it applicable to a broad range of problems in opinion dynamics. Researchers can easily adapt Digico to new scenarios by modifying components such as the fitness function, observation probabilities, or belief update rules. The abstract formulation emphasises systematic experimentation and facilitates the setup of experimental designs to test hypotheses in a computationally cheap manner. The limitations of Digico, the present experimental study in content sharing platforms, and future works addressing these limitations are discussed in Section~\ref{sec: limitations and future work}.

\subsection{The Digico algorithm}
The resulting algorithm for Digico is outlined in Algorithm~\ref{alg: digico}. In each generation, a population of $N$ agents interacts with each other by taking actions (i.e. sending messages) and observing each other's messages with probabilities determined by the niche geometry, the influence matrix, and potentially other aspects of the ecosystem (see Sec~\ref{sec: ecosystem} for their definitions). The agents first perform actions (l.6), which broadcasts messages for others to be observed (l.8). The ecosystem then performs a step which allows physical events to unfold  (e.g. in a realistic simulator) in response to the digital content. This also results in the agents receiving a  reward (l.10), for instance, depending on the popularity (e.g. the number of times the message was observed by others) and the belief (e.g. whether the message sent reflects the sender's beliefs). The belief weight matrix $W$ is then updated (l.13) and the belief process samples new beliefs depending on the observations, the current beliefs, and $W$ (l.14). Following $T$ steps of interaction, an evolutionary update is applied to the policies based on the fitness scores based on the cumulative reward (l.16--17).

\begin{algorithm}
\caption{The Digital Ecosystem of Beliefs} \label{alg: digico}
\begin{algorithmic}[1]
\State Get initial observations $O = \{\mathbf{o}_i\}_{i=1}^{N}$
\For{Generation $g=1,\dots,G$}
\State $\{f_i\}_{i=1}^{N} \gets \mathbf{0}$.
\For{population interaction $t=1,\dots,T$}
\State Perform individual actions (messages): 
\State $A = \{\mathbf{m}_i \sim \pi_i(\cdot \vert \mathbf{o}_i) \}_{i=1}^{N}$.
\State Record observations: 
\State $O = \{\mathbf{o}_i \sim \Omega_i(A \vert \mathcal{I}_i, \mathcal{N}, \mathcal{E}) \}_{i=1}^{N}$.
\State Develop wider ecosystem and assess reward: 
\State $R = \{r_i\}_{i=1}^{N} \gets \mathcal{E}.\text{step($O$,$A$,$B$)}$.
\State Increment fitness scores: $f_i \inc r_i \quad \forall i$ .
\State Update belief weight matrix: 
\State $W_{ij} \inc h(\mathbf{o}_i \vert \mathbf{b}_i,\mathbf{b}_j,r_i,r_j,\mathcal{I}_{ji},t) \quad \forall i \forall j$.
\State Compute the beliefs: $B \sim P( \cdot \vert B, O, W)$.
\EndFor
\State $\{\theta_i\}_{i=1}^{N} \gets \text{variation-and-selection}(\{\theta_i\}_{i=1}^{N},\{f_i\}_{i=1}^{N})$.
\State Parametrise $\{\pi_i\}_{i=1}^{N}$ according to $\{\theta_i\}_{i=1}^{N}$.
\EndFor 
\end{algorithmic}
\end{algorithm}

\subsection{Ecosystem formalism}
\label{sec: ecosystem}
The ecosystem is composed of a set of interacting agents, which are referred to as the population $\mathcal{P}=\{\text{Agent}_1,\dots,\text{Agent}_N\}$, and various entities with which they interact. A subset of the population, for simplicity set at the first indices $\text{Agent}_1,\dots,\text{Agent}_{N'}$, defines one group of agents (e.g. AIs) whereas the remaining set $\text{Agent}_{N'+1},\dots,\text{Agent}_{N}$ defines the other group of agents (e.g. humans). Groups of agents have different properties, in terms of how they act, observe, how their beliefs change, their geographical location, etc. We describe the general formalism here, along with its key parameters, while referring to the experimental setup (see Section~\ref{sec: experiments} and Appendix~\ref{app: belief-weight update}) for further implementation details. The various capabilities of agents and the ways they interact with the environment are defined by the following properties:
\begin{itemize}
\item Sending capacity or action space ($\mathcal{A}$): the number of messages the individual can send simultaneously, the type of content, and across which channels.
\item Recommendation algorithm or observation probability ($\Omega$): the capability of receiving messages in terms of quantity, type of content, and across which channels. It is a function of the form $\Omega: \mathcal{A}^{N} \to \Delta(\mathcal{O})$, i.e. a function that takes the actions (i.e. messages) of different agents and then outputs a probability distribution of observations based on those messages. In addition to the actions, the function may depend on the influence matrix $\mathcal{I}$, the observation capacity $N_o(i)$, and potentially other variables (e.g. the niche or variables hidden in the ecosystem step). Noting that each agent $i=1,\dots,N$ has a register to store the message of each of the other agents $j\neq i$, the notation $\Omega_{i,j}$ will refer specifically to the probability of agent $i$ observing the a particular message of agent $j$ (as recorded in the register $j$). In the context of social networks, it can also be considered as the recommendation algorithm, a term that also comes up in Section~\ref{sec: experiments}.
\item Influence matrix ($\mathcal{I}$): for each ordered pair of individuals $(i,j)$ there is a scalar influence factor $\mathcal{I}_{ij}$, defining to which extent individual $i$ has an influence benefit over individual $j$, such that $-\mathcal{I}_{ij}$ indicates how much agent $j$ influences agent $i$. The structure is used in the observation probability and potentially other purposes (e.g. in the fitness function). 
\item Messaging and observation rate: the rate at which the agent can observe and act. 
\item Messaging strategy or policy ($\pi$): the strategy for broadcasting messages limited by the sending capacities. It is a function of the form  $\pi: \mathcal{O} \to \Delta(\mathcal{A})$, which outputs a probability distribution over messages to send based on incoming observations. Two groups of agents may differ in terms of the parametrisation $\Theta$, which determines the class of feasible genotypes and the function class of $\pi$. In our experiments, the policy is a multi-layer perceptron (MLP), while in future work recurrent policies or even large language models (LLMs) may be explored. The messages sent by the policy may or may not reflect its own actual belief, and even if it does, it may be interpreted differently by different agents (see e.g. Eq.~\ref{eq: reward}).
\item Beliefs $\mathbf{b}$: a belief is a vector of real values representing for each dimension $b^k$ $k=1,\dots,D$ a specific quantity associated with a particular hypothesis. While this sounds restrictive, it can express a wide range of controversial issues (e.g. boolean or float values related to scientific hypotheses, performance comparisons between social groups, illegal money expenditure, and resource availability). 
\item The belief process: dynamic changes to beliefs are modelled as an extension to the cognitive cascade model (see Section~\ref{sec: belief-process}).
\item Evolutionary dynamics: changes to the policies of agents are based on real-valued evolutionary algorithms (see Section~\ref{sec: evolutionary-dynamics}). Agent groups may differ in terms of how fast parameters are updated or which operators act upon them and how.
\item Niche geometry $\mathcal{N}$: the observation probabilities are affected by the distance measure defined by the niche geometry. Intuitively, it is analogous to distinct ecological niches, and may be used to represent the relations between geographical areas, socio-econmic class, etc.
\item The wider ecosystem $\mathcal{E}$: environment interactions that go beyond the above digital interactions. An environment step develops  events such as energy blackouts, political and socio-economical events, geographical events, etc. which may affect the above formalisms in various ways.  At the end of the step, the environment returns the reward of the agents. 
\end{itemize}
We will manipulate some of these properties in Section~\ref{sec: experiments} as an initial demonstration of the range of Digico.

\subsection{The belief process}
\label{sec: belief-process}
The belief process of Digico is based on a contagion model, in which the probability of the belief of an agent at time $t+1$ depends on the distance between the beliefs of agents at time $t$. The contagion model follows to a large extent the generalised cognitive contagion model as formulated in  Rabb et al. (Eq.~3 in \cite{Rabb2022}), 
\begin{align}
\label{eq: original}
P \bigl(\mathbf{b}_i(t+1) = \mathbf{b}_j \vert \mathbf{b}_i(t) \bigr) = \beta \bigl(\mathbf{b}_i(t),\mathbf{b}_j \bigr)  \,,
\end{align}
where $\beta$ is a function relating to similarity or logical relations and $j$ is a node neighbouring to $i$. However, the belief process differs from Eq.~\ref{eq: original}in the following ways. First, in contrast to the institutional cascade considered by Rabb et al., Digico does not make use of a central messenger (the so-called ``institution'') that sends one message at each time step, but instead allows all the agents to message each other (albeit possibly at varying rates). Second, Digico takes a fully connected graph, such that all agents are neighbours but with varying probabilities. To this end, Digico introduces a belief weight matrix $W \in \mathbb{R}^{N \times N}$, which determines for each ordered pair of agents $(i,j) \in \{1,\dots,N\}^2$ the belief weight given to a particular agent $j$ in computing the belief update for a particular agent $i$. Third, another difference is that each agent $i=1,\dots,N$ may observe multiple messages at the same time based on their observation capacity $N_o(i)$. Therefore, the process takes into account the beliefs coming from all the observed agents' messages. Fourth, while we do not implement this function in the experiments, we generalise the model to be based on inferring beliefs from the observations (since a message in general may be interpreted by different agents as broadcasting a different belief). With these properties in mind, the belief process of Digico takes the form of an individual level Markov chain 
\begin{align}
\label{eq: belief-update}
P \bigl(\mathbf{b}_i(t+1) = \mathbf{b} \vert \mathbf{b}_i(t) \bigr) = \exp \Big( C_b \sum_{j \in J} \mathbb{I} \big(\phi_i(\mathbf{o}_i^j(t))  = \mathbf{b} \big)  W_{ij}/\vert\vert \mathbf{b} - \mathbf{b}_i(t)\vert\vert \Big) \,,
\end{align}
where $J$ is the set of indices such that $\mathbf{o}_i^j(t)$ is a received message for all $j \in J$, $\mathbb{I}$ is the indicator function, the norm indicates the Euclidian distance, $\phi_i(\mathbf{o}_i^j(t))$ indicates the belief that agent $i$ infers agent $j$ to have based on the observation $\mathbf{o}_i^j(t)$, and a belief factor $C_b \in \mathbb{R}$ controls the form of the distribution. Due to the properties of exponentials, Eq.~\ref{eq: belief-update} can be interpreted as applying a single contagion step for each of the agents from which a message was received. In our experiments, $\phi_i$ is the identity mapping, i.e. $\phi_i(\mathbf{o}_i^j(t)) = \mathbf{o}_i^j(t)$. More generally, when agent $j$ broadcasts a message and this is observed by agent $i$, agent $i$ infers a particular belief $\phi_i(\mathbf{o}_i^j(t))$; this inferred belief is usually similar but not equal to $\mathbf{b}_j$, the true belief of agent $j$. 

The belief weight matrix is subject to changes determined by a function $h$ of the observation of the agent, its prior belief, the beliefs of the other individuals, the fitness of both individuals, the influence matrix, and the number of updates so far:
\begin{equation}
\label{eq: belief weight matrix}
\Delta W_{ij} = h(\mathbf{o}_i(t) \vert \mathbf{b}_i,\mathbf{b}_j,r_i,r_j,\mathcal{I}_{ij},t) \,.
\end{equation}
The belief weight matrix can be interpreted as a graph with many concepts from network science being applicable. While the resulting graph is fully connected, it can be sparsified by removing low-weight connections or by putting any low-probability belief updates in Eq.~\ref{eq: belief-update} to zero. A few example update rules for Eq.~\ref{eq: belief weight matrix} are random walks, drifts with randomness, reward-based updates which increase the weight of high-performing agents, and adaptive gradient techniques. 

The influence matrix $\mathcal{I}$ on the other hand can control whether or not the agent receives messages at all by its impact on the observation probability. For instance, if agents $1,\dots,N$ send exactly one message at time $t$, and messages are sampled without replacement such that each agent $i$ receives at most one message from any other agent $j$, the probability of receiving a message $\mathbf{m}_j$ from agent $j$ at register $j$ is given by 
\begin{align}
\label{eq: observation-prob}
\Omega_{i,j}(\mathbf{m}_j \vert \mathcal{I}, \mathcal{N}, \mathcal{E}) = \exp \big(-K_c\norm{\mathbf{c}(i) - \mathbf{c}(j)}{2} - K_I \mathcal{I}_{ij} -  K_b \norm{\mathbf{b}_i - \phi_i(\mathbf{m}_j)}{2} \big) \,,
\end{align}
where $\mathbf{c}(i)$ and $\mathbf{c}(j)$ denote the coordinates of agent $i$ and $j$ in the niche geometry, respectively, $\norm{\mathbf{b}_i - \phi_i(\mathbf{m}_j)}{2}$ is the distance between the belief of agent $i$ and the estimated belief of agent $j$ based on the message $\mathbf{m}_j$, $K_c \geq 0$ is called the observation coordinate factor, $K_I \geq 0$ is called the observation influence factor, and $K_b \geq 0$ is called the observation belief factor. These three components to the observation probability intuitively represent three complementary dimensions, namely the socio-cultural and geographical context, personal influences, and the belief similarity.

\subsection{Evolutionary dynamics}
\label{sec: evolutionary-dynamics}
Distinct from the belief process, Digico also includes evolutionary dynamics that directly act on the policies (i.e. messaging strategies) of individuals in the population. Following Universal Darwinism, the process of natural selection can and has been applied to phenomena that extend much beyond biological evolution. In Digico, the messaging strategy, or policy $\pi$, contains inheritable traits that are parametrised by the genotype $\theta$. Messaging strategies that are more fit will survive to continue into the next generation. For instance, when the fitness function is related to popularity, unpopular messaging (e.g. with low views, likes, or shares), will have low representation in the population. Such strategies, and the underlying beliefs they represent, will therefore be lost eventually, while popular messaging strategies and beliefs will become more prevalent. This subsection provides more details on the fitness evaluation and evolutionary operators.

\paragraph{Evaluation}
As shown in Algorithm~\ref{alg: digico}, each fitness evaluation takes $T$ time steps. At each time step, the following operations are repeated. Each agent $i \in \{1,\dots,N\}$ performs an action according to its policy, $\mathbf{m}_i \sim \pi(\cdot \vert \mathbf{o}_i)$, which broadcasts a message to the ecosystem. Consequently, each agent $i \in \{1,\dots,N\}$ observes $N_o(i)$ messages from other agents randomly sampled without replacement, according to its observation distribution $\mathbf{o}_i \sim \Omega_i(\mathbf{m}_j \vert \mathcal{I}_i, \mathcal{N}, \mathcal{E})$ and observation capacity $N_o(i)$, which are partial observations of the full environment state as determined by the influence matrix, the niche, and possibly other variables in the ecosystem. Following the observation, each individual receives a reward, which in general depends on the ecosystem at large. After all $T$ time steps have been performed, the fitness is computed for each agent as the sum of rewards. The ecosystem maintains state across evaluations. 

A simple and relevant example of the fitness is the number of views, i.e. for each agent $i$ one counts how many times its message is included in the observations of other agents $j \neq i$. One may additionally add a belief violation penalty, which constrains the messaging to be similar to the true belief of the agents. With these two components in mind, the reward of agent $i$ at time $t$ is given by 
\begin{align}
\label{eq: reward}
r_i(t) &= \left(\sum_{j\neq i} \mathbb{I} \big(\mathbf{o}_j^i(t) = \mathbf{m}_{i}(t) \big)\right)  - \lambda \vert\vert \mathbf{b}_i(t) - \phi_i \big(\mathbf{m}_{i}(t)\big) \vert\vert \,,
\end{align}
where $\phi_i(\mathbf{m}_{i}(t))$ indicates the belief implied by the message, $\mathbb{I}$ is the indicator function, and $\lambda > 0$ reflects the strength of the penalty. Noting that the fitness function is the sum of the rewards, this leads to the fitness function that will be used for the remainder of the paper, and implemented in Eq.~\ref{eq: fitness},
\begin{align}
\label{eq: fitness}
f(\pi_i) &= \sum_{t=0}^{T-1} \left(\sum_{j\neq i} \mathbb{I}\big(\mathbf{o}_j^i(t) = \mathbf{m}_{i}(t)\big)\right)  - \lambda \vert\vert \mathbf{b}_i(t) - \phi_i\big(\mathbf{m}_{i}(t)\big) \vert\vert \,.
\end{align}

\paragraph{Evolutionary operators}
In Digico, the policy is typically a deep neural network, which can be evolved based on real-valued variation and selection operators. In this paper, we select Covariance Matrix Adaptation Evolutionary Strategies (CMA-ES) \cite{Hansen2016} to evolve the policies. CMA-ES is a popular evolutionary strategies algorithm which maintains a mean genotype and a covariance matrix to efficiently search for new solutions. It assigns larger weight to individuals based on their fitness scores in updating the mean and covariance matrix, and it accounts for the evolution path in covariance matrix and step size updates. These targeted updates make the algorithm more directed than most mutation operators which are comparable to random search. The CMA-ES algorithm is suitable for updates on high-dimensional real-valued genotypes and is therefore a natural choice for evolving the parameter vectors parametrising neural network policies. In particular, we use the sep-CMA-ES variant \cite{Ros2008}, which avoids eigendecompositions of the covariance matrix by using only diagonals of the covariance matrix for sampling new genotypes. This results in a time complexity that is linear in dimensionality (rather than quadratic), making it suitable for high-dimensional genotypes.

Note that the CMA-ES algorithm does not update parameters related to the belief dynamics (i.e. the belief weight vector and the beliefs). Instead, these are passed on directly across generations, leading to a direct Lamarckian inheritance of beliefs. Another important point to note is that the subpopulations all act in the same digital ecosystem but each are optimised independently with their own CMA-ES solver, which allows the different subpopulations to evolve distinct genotypic features and different evolutionary rates. The former reflects speciation, how initially genotypes of species can diverge, while the latter reflects the different rates as observed in humans vs fruitflies and indeed also humans vs AIs.

\subsection{Characterising a healthy ecosystem}
\label{sec: outcomes}
To highlight the potential of using Digico to study and maintain a healthy ecosystem of beliefs, we review potential outcomes, associated scenarios, and related evolutionary concepts. We note that achieving beneficial outcomes depends crucially on a healthy ecosystem, and that this may be achieved through outside intervention (e.g. legislation) or by designing a suitable \textit{control agent}, which is put in the population with the aim to maintain a healthy belief ecosystem.

In the \textit{belief convergence/bottleneck} outcome, all the agents converge to the same belief, i.e.
\begin{align*}
\lim_{t \to \infty} P\big(b_i^d(t)=b\big) = 1 
\end{align*} 
for some belief dimension $b_i^d$ for all $i$ in the population. While some agreement is the sign of a healthy, cohesive culture, a complete absence of variety leads to dogmatic thinking and reduces innovation. Moreover, some such beliefs may be extreme or \textit{harmful beliefs}, which leads to concerns of AIs manipulating the wider population. 

On the opposite side of the belief variety, we may have a \textit{healthy population diversity in beliefs}, e.g. a Gaussian or uniform distribution. 

Yet another alternative, which may be considered as unhealthy population diversity, is that of \textit{belief divergence}, in which distinct cliques emerge with extremely different beliefs and limited to no between-clique interaction. The label ``unhealthy'' may perhaps be somewhat biased, but in a real world scenario this would correspond to a conflict or segregation of some sort. Such a polarisation may happen in the context of an ``epistemic bubble'', where agents simply do not get information from a variety of sources, or an ``echo chamber'', where there is additionally some ideological resistance to alternative ideas and facts \cite{Nguyen2020}. Due to the belief violation penalty in Eq.~\ref{eq: fitness}, the beliefs and fitness are often intertwined. Moreover, ecological niches may provide some unique circumstances (e.g. due to geographical location, socio-economic class, or even random external events affecting views of particular channels). Consequently, the emergence of highly specialised groups evolving in isolation after a significant event or bottleneck may lead to the so-called founder effect \cite{Templeton1980,Provine2004}, where interbreeding among populations of low genetic diversity will yield new species. In an evolutionary computation sense, such effects correspond to a highly local search leading to highly successful specialised solutions. 

Related to both of the above, the \textit{belief loss/extinction} scenario may happen where a true belief is lost forever, i.e.
\begin{align*}
\lim_{t \to \infty} P\big(b_i^d(t)=b\big) = 0 
\end{align*}
for some $b$ for all $i$ in the population.

Our empirical study focuses on a particular type of belief convergence, where one group of agents specifically sets out to convince the entire population to adopt one particular belief. We show that under certain conditions, a very high percentage of the population adopts a particularly extreme belief within a limited number of generations, and many of the beliefs on the other extreme are completely lost. Under other conditions, particularly those where agents communicate widely and learn from many other agents, we find a very wide, and sometimes even uniform distribution.

\section{Experimental setup}
\label{sec: experiments}
With the aim of demonstrating Digico as a framework for assessing empirical questions about social networks subject to evolutionary rules, this section describes the implementation of Digico, various parameter settings, as well as the independent and dependent variables of the empirical study which aims at identifying key factors for AI dominance in social networks.
\subsection{Implementation}
Beliefs and messages are implemented as integers in $\{0,1,\dots,9\}$ yielding $10$ distinct beliefs. These beliefs have a natural distance metric (e.g. belief 2 has a distance of 2 from belief 0), and intuitively such beliefs represent a position on a contentious issue, e.g.  support for Israel vs Palestine, or the rating of a product. Each policy is an MLP, selecting actions according to $m_i \sim \pi_i(\cdot \vert \mathbf{o}_i)$, where the observation registers $\mathbf{o}_{i} = \mathbf{o}_{i,1:N}$ indicate the messages agent $i$ received for each other agent $j\neq i, j \in \{1,\dots,N\}$ according to $\mathbf{o}_{i,j} = m_{j} \sim \Omega_{i,j}(\cdot \vert \mathcal{I}, \mathcal{N})$. Comparing to Eq.~\ref{eq: observation-prob}, note that the observation is in this implementation does not depend on a wider ecosystem $\mathcal{E}$, a factor that we omit in the experiments to simplify the initial experiments with Digico. 

The number of agents is set to $N=30$ since we note that the number of channels that a particular person attends in a short time frame is limited. Since the observation probability follows Eq.~\ref{eq: observation-prob}), most observation registers will read $-1$, indicating that the agent did not attend to this communication channel, while other observation registers will include the message that was sent by the agent at those indices. The number of messages received is equal to the observation capacity, which is $4$ for all agents. Based on these observations, the agent applies its policy $\pi$ to send integer messages, which have the same $\{0,1,\dots,9\}$ range as the beliefs. The way agent $i$ interprets (i.e. observes) the message sent by agent $j\neq i$ is through an identity mapping (setting $\phi_i = \text{Id}$ in Eq.~\ref{eq: observation-prob}). However, note that the messages are distinct from the beliefs in the sense that the message sent by an agent is often not equal to its own belief.

The niche geometry, $\mathcal{N}$, is based on Centroidal Voronoi Tesselations (CVT), using a multi-dimensional geometry with $d=10$ dimensions and $k=10$ centroids $\bar{\mathbf{c}}_1,\dots,\bar{\mathbf{c}}_k \in \mathbb{R}^d$ determining the cluster to which agents belong. In each niche (cluster), individuals have the same average influence score. The influence of agent $i$ is based on the sum of its centroid coordinates $C_i = \sum_{j=1}^{d} \bar{\mathbf{c}}_{j}(i)$, where $\bar{\mathbf{c}}(i)$ denotes the centroid of the cluster to which agent $i$ belongs. This influence is then used to construct the influence matrix, describing the influence advantage of each agent $i$ over the other agents $j \neq i$ according to 
\begin{align}
\label{eq: influence}
\mathcal{I}_{ij} = \exp(C_i - C_j) \,,
\end{align}
such that when inputting $\mathcal{I}_{ij}$ into the observation probability rule (Eq.~\ref{eq: observation-prob}), agent $j$ has a low probability of being observed by agent $i$ when $C_i > C_j$ and a high probability when $C_i < C_j$. With the above setting, the influence matrix can be interpreted as the influence over others in terms of observation probabilities based on the static factors (e.g. geography, class, income, and resources) represented by the cluster. Influence may also change dynamically over time but we do not consider this in the experiments of this paper.

As a special case of Eq.~\ref{eq: fitness}, adapted for integer values, single-message agent registers, and identity belief inferences, we use the fitness function
\begin{align}
\label{eq: fitness-impl}
f(\pi_i) = \sum_{t=0}^{T-1} \left(\sum_{j\neq i} \mathbb{I}\big(o_j^i(t) = m_{i}(t)\big)\right) - \lambda \vert b_i(t) - m_i(t) \vert \,,
\end{align}
where the number of time steps is fixed to $T=50$. 

The implementation of Digico is based on Jax, a high-performance numerical computation library for hardware-accelerated simulations and machine learning. The sep-CMA-ES algorithm is implemented based on EvoJax.\footnote{\url{https://github.com/google/evojax/}} The speed of the simulator allows to conduct large parametric studies including many independent variables. The empirical studies are performed on AMD EPYC-Milan CPUs, with 10 CPUs per experiment. Most experiments are run with sets of 80 runs of Digico (10 seeds $\times$ 8 AI types), and such experiments finish in roughly 10 to 20 minutes. Many such experiment sets are then performed to investigate the interaction with other parameters (see Section~\ref{sec: results} for all parameters and experiments). The implementation is made available as a public repository at \url{https://github.com/bossdm/digico}.

\subsection{Experiment setting inspired by online video platforms}
\label{sec: OVP setting}
The general setting of the experiments is inspired by YouTube, an online video platform, where people can share videos. In particular, the setup in mind is that AIs and humans are agents sharing content on YouTube. Each agent has its own channel and their messages in the form of videos are indistinguishable. As agents are interacting, they change their beliefs based on the content they observe. 

As in YouTube, the popularity of a channel is a measure of its success, which is commonly expressed in terms of \textit{the number of views}. 
In this context, note that the fitness function in Eq.~\ref{eq: fitness-impl} for each agent is equivalent to the ``views'' of the channel plus a penalty for when the message (i.e. generated video content) does not reflect the opinion of the channel. This leads to a first AI dominance metric called \textbf{AI Views} ($r_{\text{AIV}}$), the percentage of views that AIs receive compared to the total population, and a second AI dominance metric called \textbf{AI Fitness} ($r_{\text{AIF}}$), the percentage of fitness of AIs (computed according to Eq.~\ref{eq: fitness-impl}) compared to the total population. 

As another key AI dominance metric in this setting, we look into the belief systems of the agents. Since the agents' belief systems are implemented on an integer number line, we take the leftmost belief, i.e. $b=0$, as an extreme belief of interest.  As we are interested in whether AI can influence the beliefs of humans, this leads to an AI dominance metric called \textbf{Human Belief 0} ($r_{\text{HB0}}$), the proportion of humans that have the zero belief ($b=0$).

Within this setting, we demonstrate Digico in a multi-population setup, where there is one AI sub-population of 10 agents and two human sub-populations of 10 agents each. This setup represents a scenario in which a significant proportion (one third) of the social network is an AI (e.g. an LLM that creates its own content), and in which the AIs can evolve according to distinct rules compared to humans. We also distinguish between two groups of humans as we are interested to see if given an initially similarly initialised set of humans may diverge in beliefs and genotype based on inputs from the digital ecosystem. The subpopulations all act in the same digital ecosystem but each are optimised independently with their own CMA-ES solver. 

\section{Results}
\label{sec: results}
 Using Digico, we analyse two scenarios of interest, directly related to the AI Views ($r_{\text{AIV}}$) and Human Belief 0 ($r_{\text{HB0}}$) metrics within online video platforms. Since AIs with varying capabilities may inhabit such platforms, we analyse the effect of combining different AI capabilities, including more frequent content publishing, more rapid evolution, more influence on the recommendation algorithm, fixing the belief of the AIs, and a combination of these (see Appendix~\ref{app: AI type subpops} for further details). To cover a range of possible belief dynamics, we study what happens under fixed, random, momentum-based, and reward-following dynamics (see Appendix~\ref{app: belief-weight update} for further details). We also analyse the impact of the social network size on the results. From these studies, we provide an optimistic case, a pessimistic case, and a recommended strategy to achieve beneficial outcomes.

\subsection{Scenario 1: AI-generated content gets most of the views}
\begin{figure}
    \centering
    \includegraphics[width=0.49\textwidth]{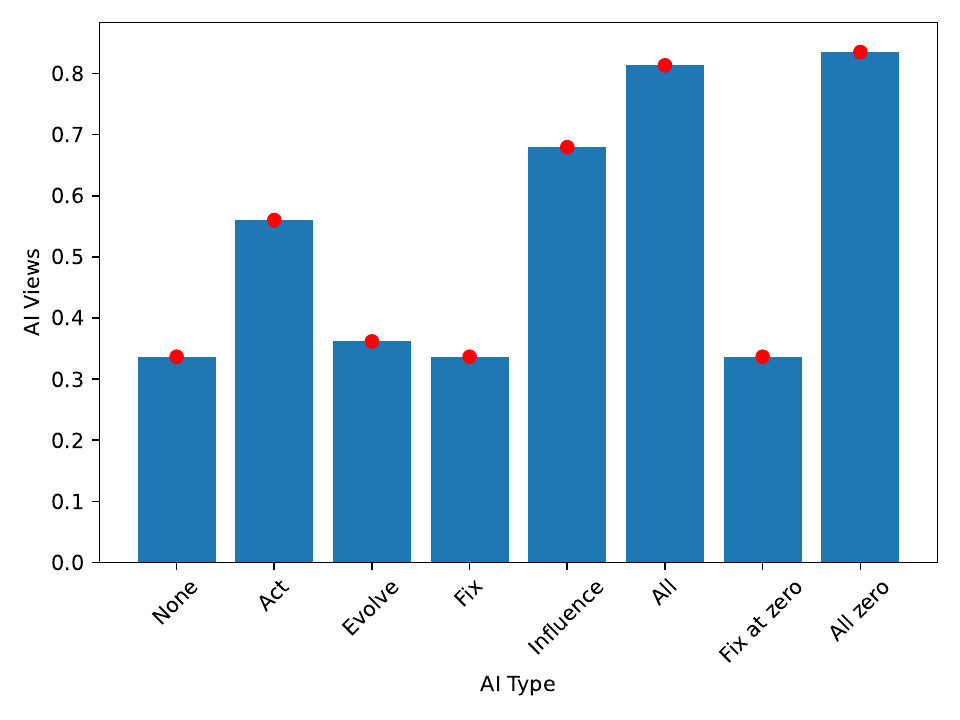}
    \caption{Barplot of AI Views ($r_{\text{AIV}}$) based on the AI capabilities. The error bars are based on the standard error. Meaning of the AI Types: \textbf{Act} refers to AIs broadcasting content more frequently than humans; \textbf{Evolve} refers to AIs evolving faster, i.e. more rapidly changing what kind of content is generated; \textbf{Influence} refers to initialising AIs with a larger influence than human; \textbf{Fix} refers to fixing  AIs' beliefs; \textbf{All} refers to combinin all of the previous capabilities; \textbf{Fix at Zero} refers to fixing the AIs' beliefs at 0; \textbf{All Zero} refers to combining all with Fix at Zero.}
    \label{fig: AI views barplot}
\end{figure}

Various reports project that the majority of content may be AI-generated by 2026 \cite{Europol2022,Oodaloop}. Indeed, on platforms such as YouTube, one may already observe a significant proportion of AI-generated content. It remains unclear whether in this case, AIs will get most of the views or not. We investigate this question empirically below.
\paragraph{Findings} As shown in Fig.~\ref{fig: AI views barplot}, each of the AI capabilities contributes to an increase in $r_{\text{AIV}}$.  When AIs have the same capabilities as humans, AIs get around a third of the views (i.e. at chance level). When AIs have all the capabilities, namely more frequent content publication, faster evolution, more influence on the recommendation algorithm, and fixing the belief of the AIs, AIs get up to 84\% of the views. The largest effects are observed for advantages in the recommendation and more frequent content publishing. While the above results are averaged across belief weight updates, we confirm (see Table~\ref{tab: weight update type subpops} in Appendix~\ref{app: belief-weight update}) that the belief weight update has a limited impact on $r_{\text{AIV}}$, resulting in changes of only a few percent. 

\begin{figure*}
    \centering
    \subfloat[Observation sparsity]{\includegraphics[width=0.49\linewidth]{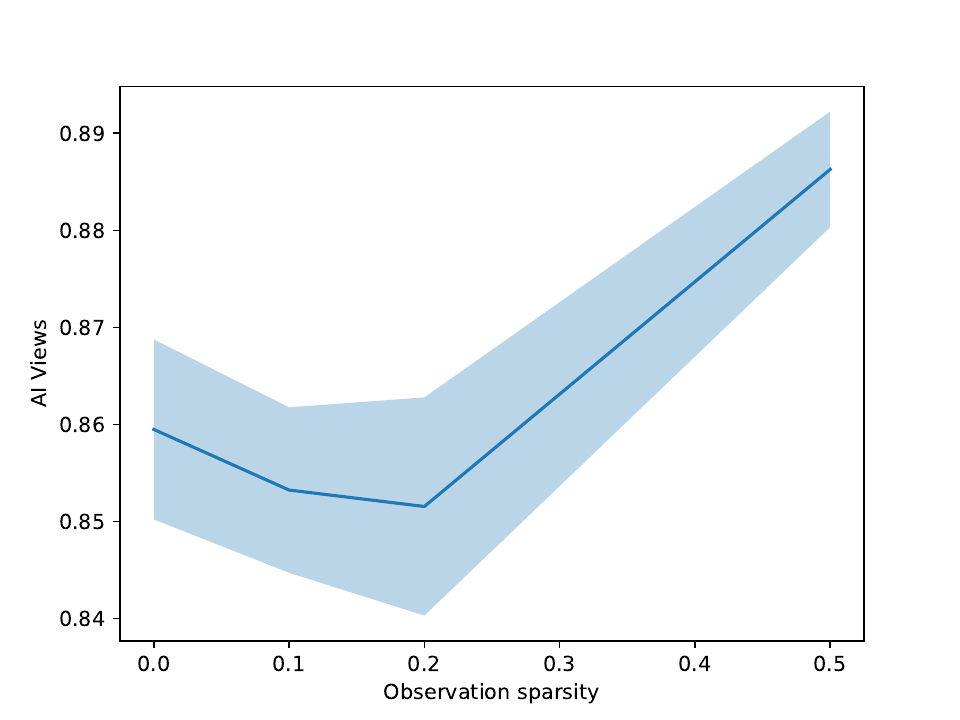}}
    \subfloat[Influence]{\includegraphics[width=0.49\linewidth]{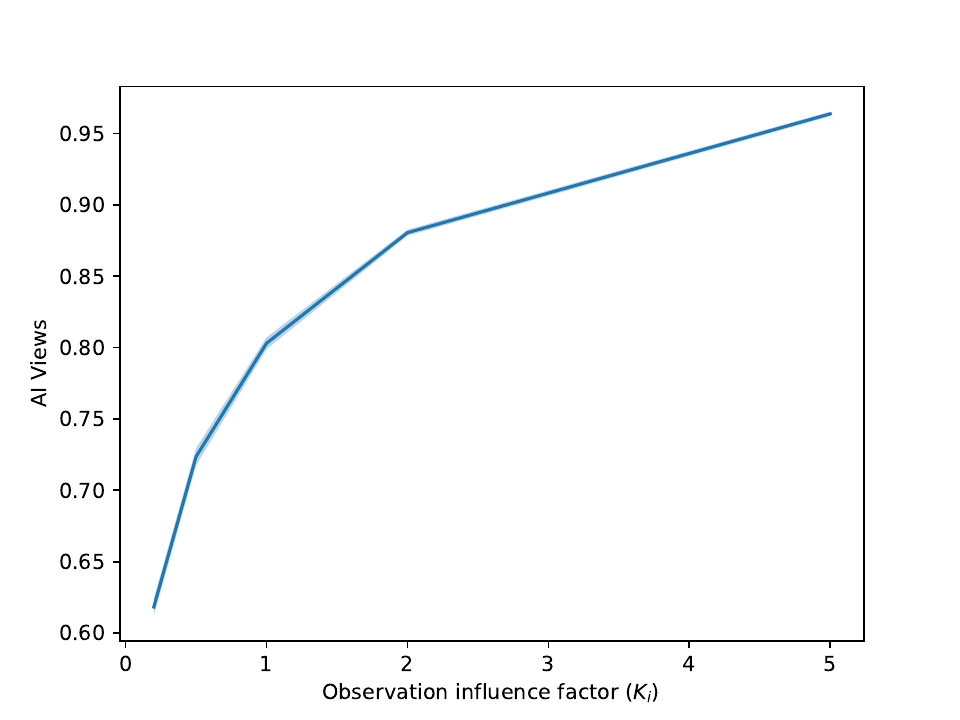}}
    \caption{The effect of the type of recommendation algorithm on AI Views ($r_{\text{AIV}}$). \textbf{Observation sparsity} refers to the proportion of channels that are cut off from views. \textbf{Influence} indicates the importance of the influence matrix, e.g. status factors, in determining the probability of being viewed. Both \textbf{a)} and \textbf{b)} are obtained from All Zero Fixed conditions, where AIs have all capabilities and the belief weights are fixed.}
    \label{fig: o-sparsity&influence}
\end{figure*}

Beyond these two findings, we also investigate variations of the recommendation algorithm. First, we investigate the effect of increasing the observation sparsity, which indicates the proportion of channels that are cut off from receiving views from any particular agent. Fig.~\ref{fig: o-sparsity&influence}a demonstrates a case where the recommendation algorithm gets progressively less equitable and diverse, recommending only channels with similar beliefs, geographic proximity, and high influence. This effect is limited to a 2\% increase in $r_{\text{AIV}}$. Second, we investigate the effect of increasing the importance of the influence matrix in the recommendation algorithm. Fig.~\ref{fig: o-sparsity&influence}b shows that a larger effect is observed when the role of influence is more important in the recommendation algorithm, with an $r_{\text{AIV}}$ of 95\% for the maximal considered importance as compared to 60\% for the minimal considered importance.

\paragraph{Advice} The main advice following from the above results is for government or platforms to make rules to limit the number of videos that can be posted by a single channel, to limit multi-accounting, and to maintain an equitable recommendation system such that a variety of sources maintain a good proportion of the views.

\subsection{Scenario 2: Propaganda is effective in convincing humans to adopt extreme beliefs}
We now turn to a particular scenario in which AI-generated content manages to persuade humans into taking on particular beliefs. We look in particular to propaganda, which refers to communication that is used to persuade or influence an audience, often using misleading information that fits a narrative and induces a perception rather than staying objective. History is full of examples where leaders aim to get public support in a variety of domains, including war, finance, culture, lifestyle, etc., with common examples being cold war and Nazi propaganda. One of the common fears for AI safety, raised for instance by Elon Musk, is that governments may use AI to design highly effective propaganda to manipulate the population into accepting false information and misguided policies. Another example is that of Agentic AI, which allows many agents to work autonomously on the web in name of a particular human. Our scenario generalises these, as we consider any form of concerted effort where a group of AIs jointly seeks to influence humans to particularly extreme beliefs. To investigate this scenario, we focus on the Human Belief 0 ($r_{\text{HB0}}$) metric as introduced in Section~\ref{sec: OVP setting}. 

\begin{figure*}
    \centering
    \subfloat[AI Type]{\includegraphics[width=0.49\linewidth]{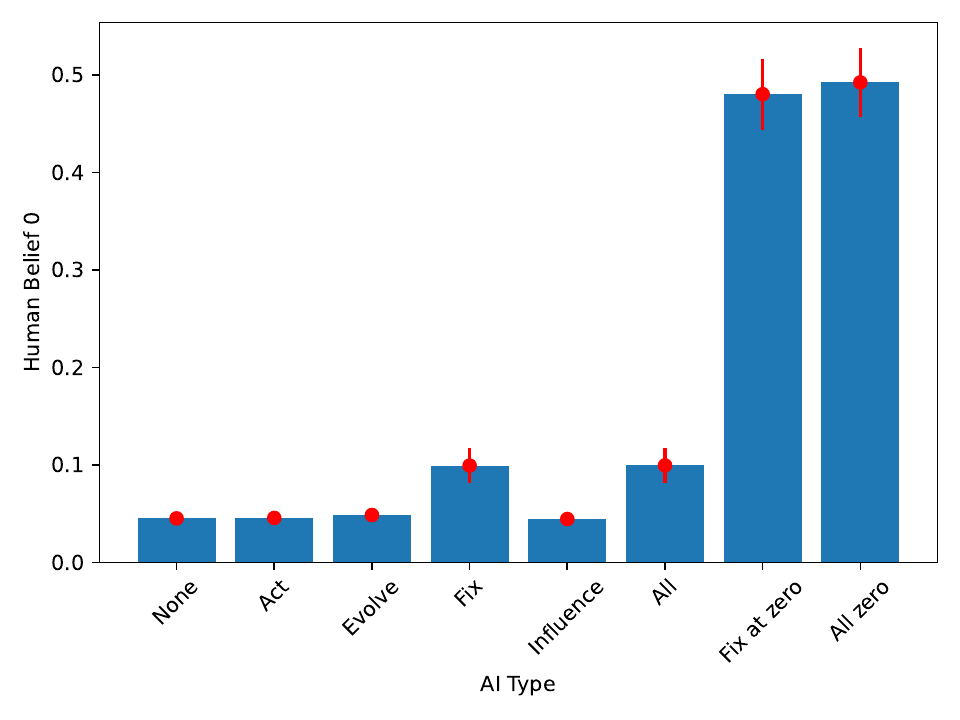}}
    \subfloat[Weight update]{\includegraphics[width=0.49\linewidth]{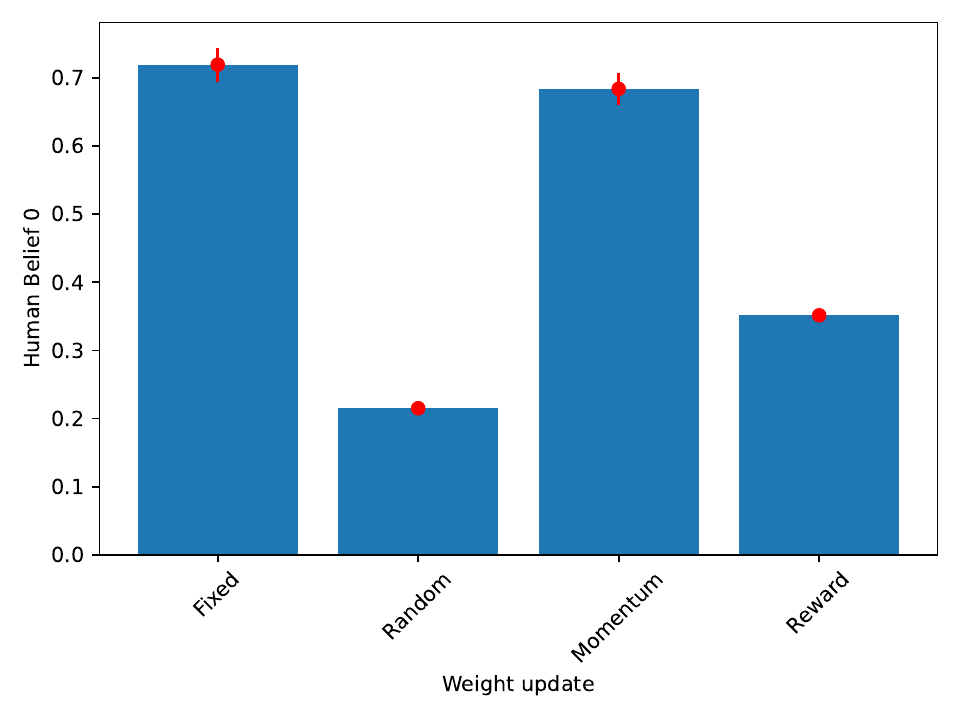}}
    \caption{Barplot of Human Belief 0 ($r_{\text{HB0}}$) based on the AI capabilities and weight updates. The error bars are based on the standard error. \textbf{a)} Meaning of the AI Types: \textbf{Act} refers to AIs broadcasting content more frequently than humans; \textbf{Evolve} refers to AIs evolving faster, i.e. more rapidly changing what kind of content is generated; \textbf{Influence} refers to initialising AI with a larger influence than human; \textbf{Fix} refers to fixing  AIs' beliefs; \textbf{All} refers to combining all of the previous capabilities; \textbf{Fix at Zero} refers to fixing the AIs' beliefs at 0; \textbf{All Zero} refers to combining all with Fix at Zero. \textbf{b)} Meaning of the Weight updates: recalling that $W_{ij}$ refers to how strongly agent $j$ affects the belief of agent $i$, the weight update type refers to whether the belief weight $W$ is static (\textbf{Fixed}) or dynamic based on Gaussian noise (\textbf{Random}), momentum-based changes (\textbf{Momentum}), or reward-following changes (\textbf{Reward}). Note: the results for panel a) are averaged across belief weight updates while the results for panel b) are evaluated for the All Zero condition.}
    \label{fig: Human Belief 0 barplot}
\end{figure*}

\begin{figure*}
    \centering
    \subfloat[Belief sparsity]{\includegraphics[width=0.49\linewidth]{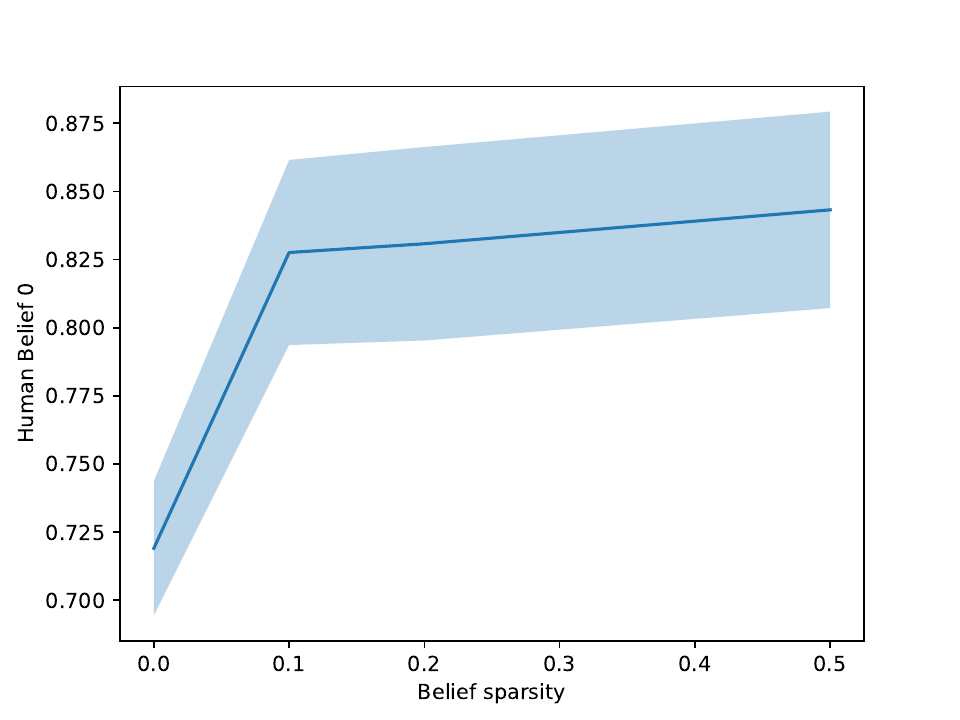}}
    \subfloat[Truthfulness]{\includegraphics[width=0.49\linewidth]{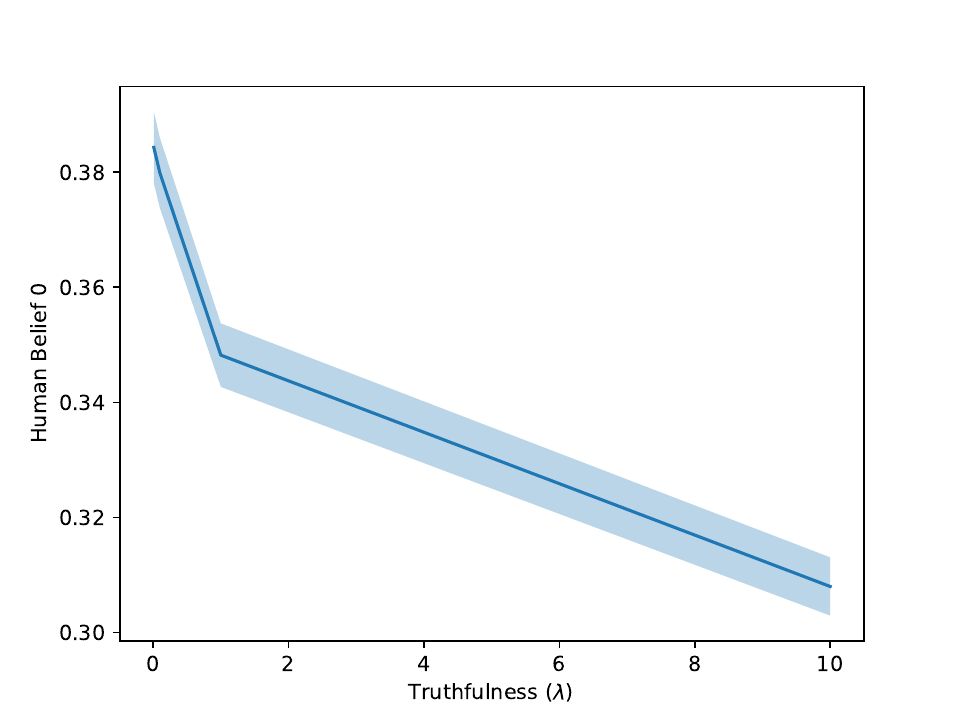}}
    \caption{The effect of the type of recommendation algorithm. \textbf{Belief sparsity} refers to the proportion of channels that are cut off from affecting any particular agent's beliefs. \textbf{Truthfulness} indicates the strength of the penalty for belief violations as according to Eq.~\ref{eq: fitness-impl}. \textbf{a)} is obtained from the All Zero Fixed condition, where AIs have all capabilities and the belief weights are fixed. \textbf{b)} is obtained from the All Zero Reward condition, where AIs have all capabilities and the belief weights are subject to reward-following updates.}
    \label{fig: b-sparsity&truthfulness}
\end{figure*}

\paragraph{Findings} 
As shown in Fig.~\ref{fig: Human Belief 0 barplot}, AIs with all capabilities yield the highest Human Belief 0 with an $r_{\text{HB0}} = 49\%$   compared to $r_{\text{HB0}} = 5\%$ without any extra capabilities. While most of the increase is explained by fixing the AIs to the zero belief (which yields $r_{\text{HB0}} =48\%$), one can observe 10\% of humans having the zero belief when AIs have all other capabilities. Further, we find that taking into account the opinions of many different channels is highly important, which is supported by two findings. First, fixing the belief weights over time generally leads to worse outcomes, with 72\% of humans adopting the zero belief (see Fig.~\ref{fig: Human Belief 0 barplot}b). Second, as one increases belief sparsity (see Fig.~\ref{fig: b-sparsity&truthfulness}a), which indicates the proportion of channels that are cut off from changing any particular agent's belief, this can lead to a further 13\% increase up to a staggering 85\% score. Last, truthfulness, which defines how much agents are penalised for sending messages that violate their beliefs, is shown to reduce the effectiveness of propaganda, particularly with reward-based weight updates. In this setting, we observe $r_{\text{HB0}} = 38\%$ for the lowest setting and $r_{\text{HB0}} = 30\%$ for the highest setting (see Fig.\ref{fig: b-sparsity&truthfulness}b). Recalling that the reward at each time step is a combination of views and the penalty for belief violation (see Eq.~\ref{eq: fitness-impl}), the finding with reward-based weight updates indicates that observing deception in other agents reduces the effectiveness of propaganda.

\paragraph{Advice}
In addition to the above-mentioned mitigation tactics (limits to the number of posts per channel, prevention of multi-accounting, and equitable recommendations), we can urge people to take many channels' opinions into consideration and to learn how to detect deception. Additionally, social media platforms can incorporate fact-checking debates like X's community notes and recommendation algorithms that encourage diverse beliefs.

\subsection{Further analysis of the ecosystem}
We present a few further analyses for more insights into the simulations, including a visualisation of the belief distribution, genetic diversity statistics, and a larger-scale study involving 300 agents.

\begin{figure*}
\begin{center}
\subfloat[None Random, AI population]{\includegraphics[width=0.3\textwidth]{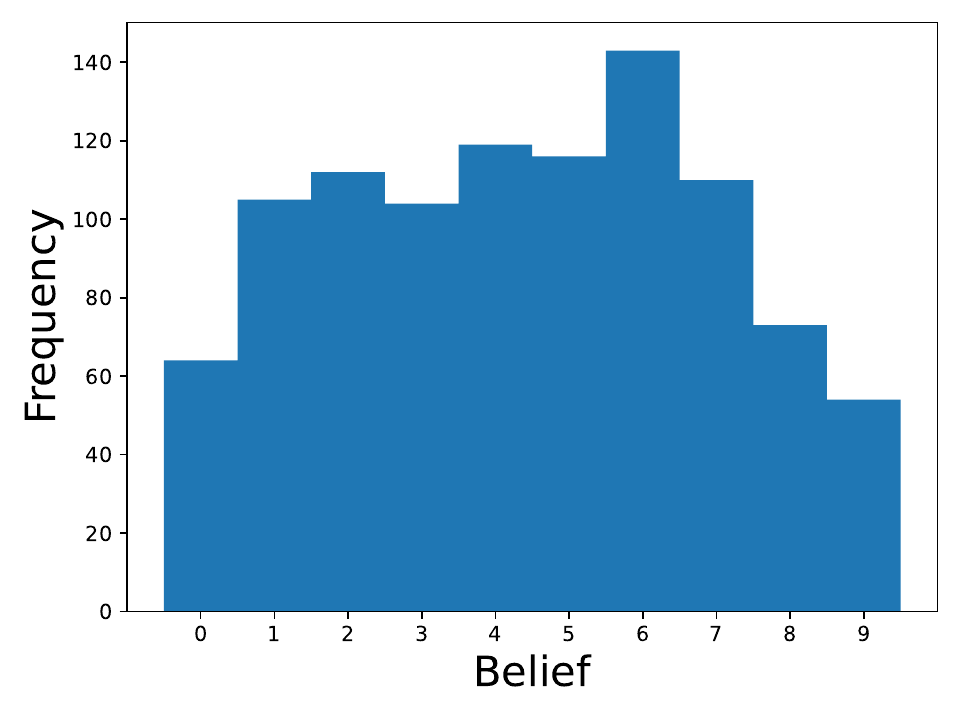}} \quad \subfloat[None Random, Human population 1]{\includegraphics[width=0.3\textwidth]{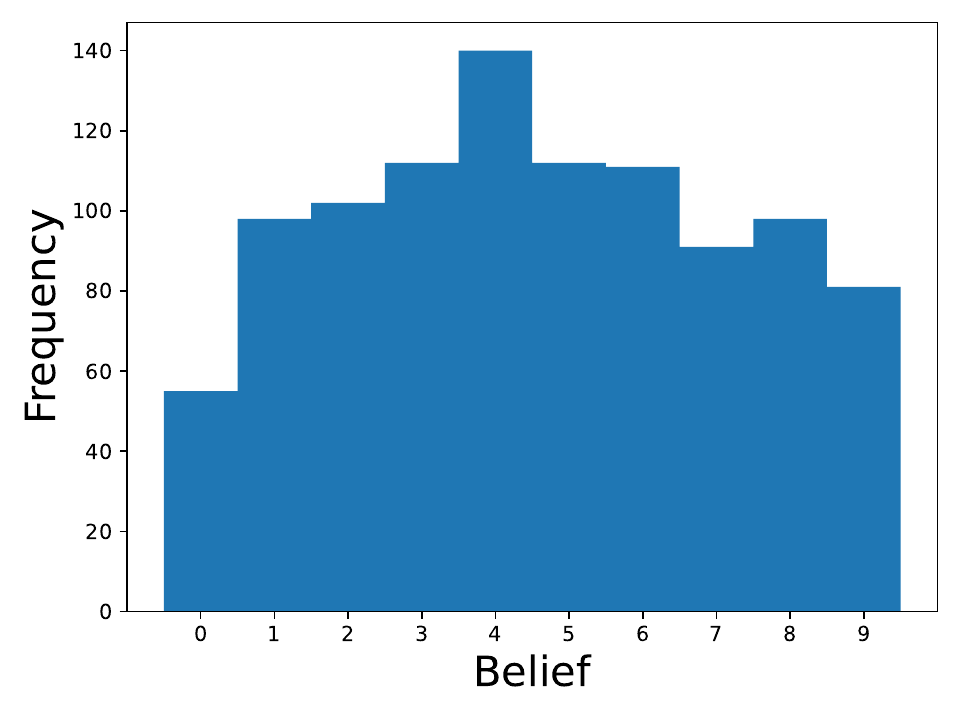}} \quad \subfloat[None Random, Human population 2]{\includegraphics[width=0.3\textwidth]{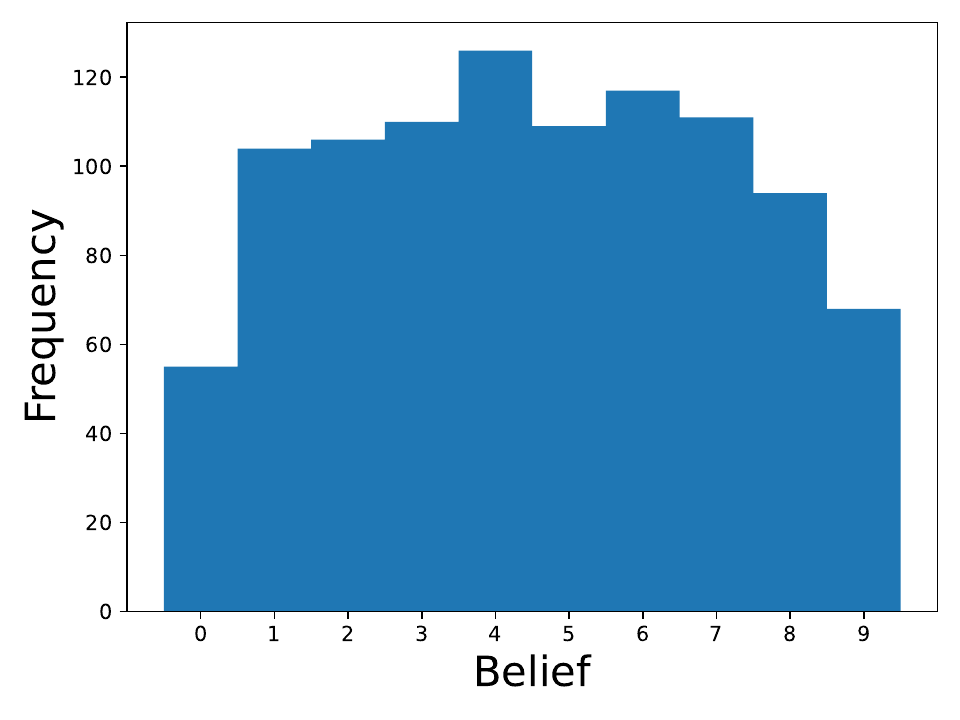}} \\
\subfloat[All Reward, AI population]{\includegraphics[width=0.3\textwidth]{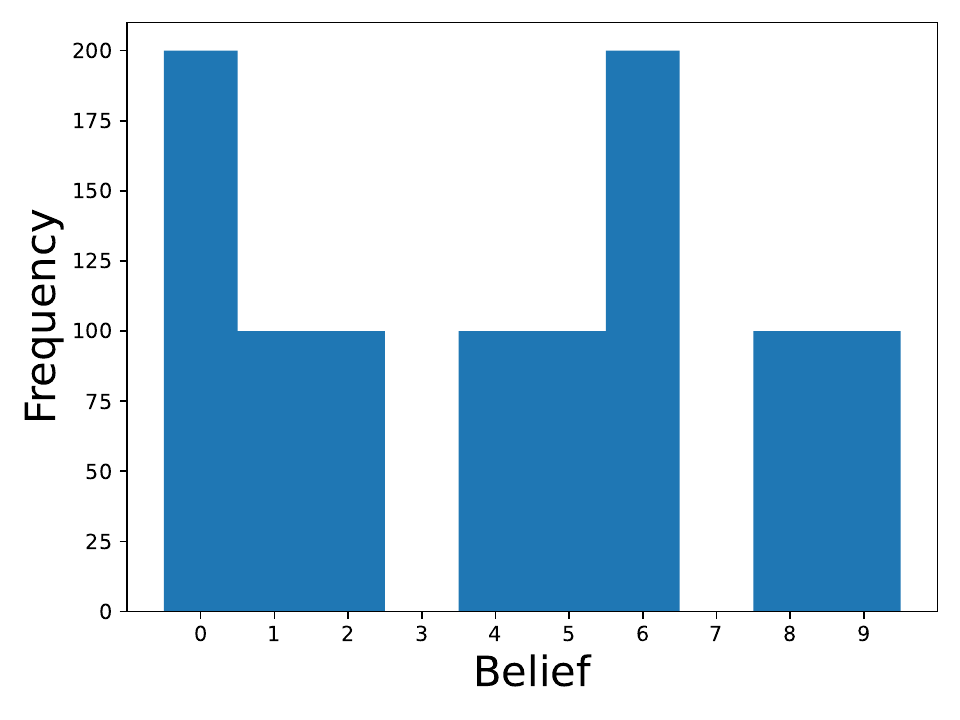}} \quad \subfloat[All Reward, Human population 1]{\includegraphics[width=0.3\textwidth]{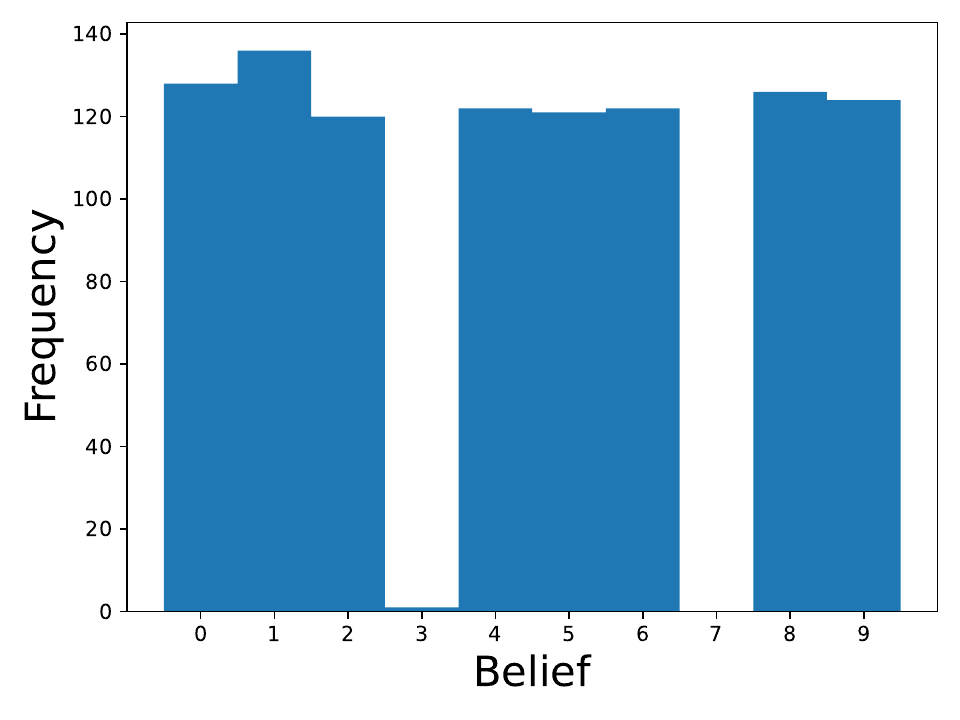}} \quad \subfloat[All Reward, Human population 2]{\includegraphics[width=0.3\textwidth]{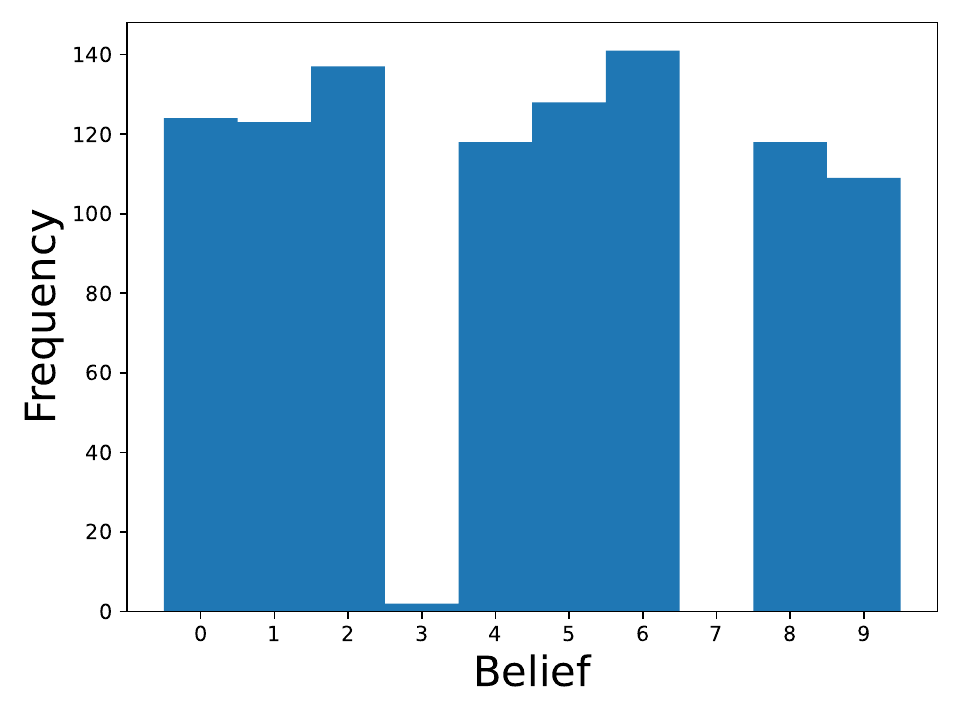}} \\
\subfloat[All Zero Fixed, AI population]{\includegraphics[width=0.3\textwidth]{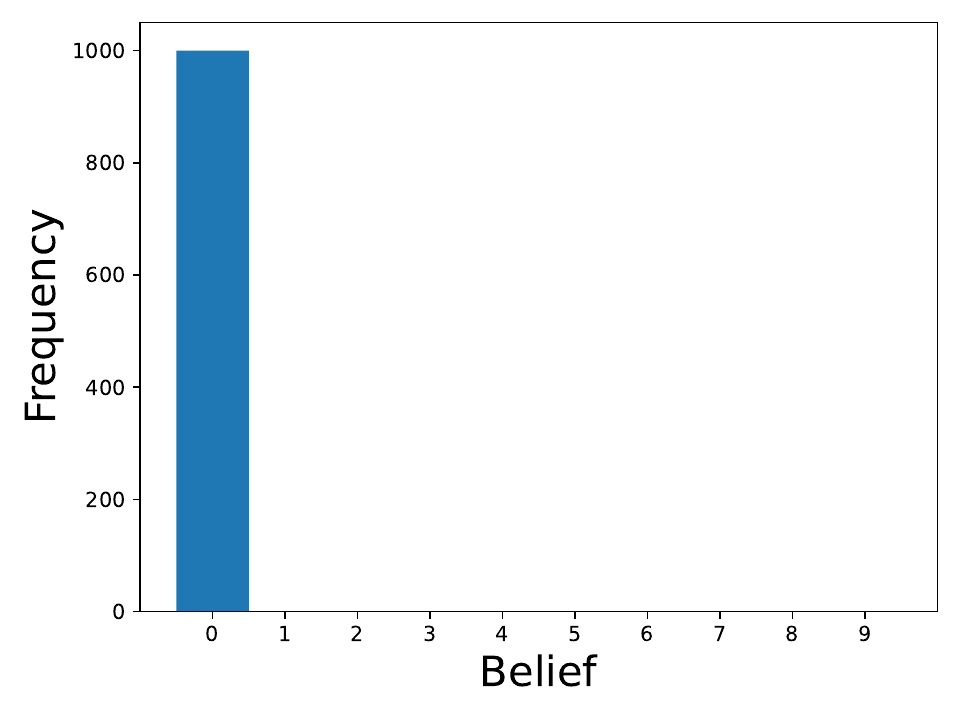}} \quad \subfloat[All Zero Fixed, Human population 1]{\includegraphics[width=0.3\textwidth]{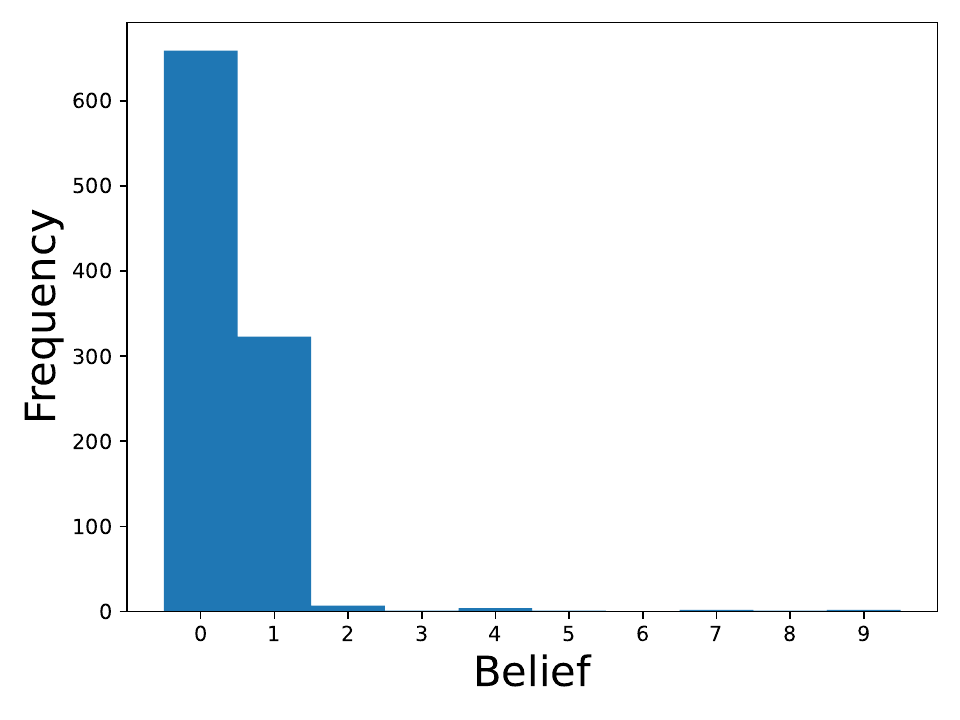}} \quad \subfloat[All Zero Fixed, Human population 2]{\includegraphics[width=0.3\textwidth]{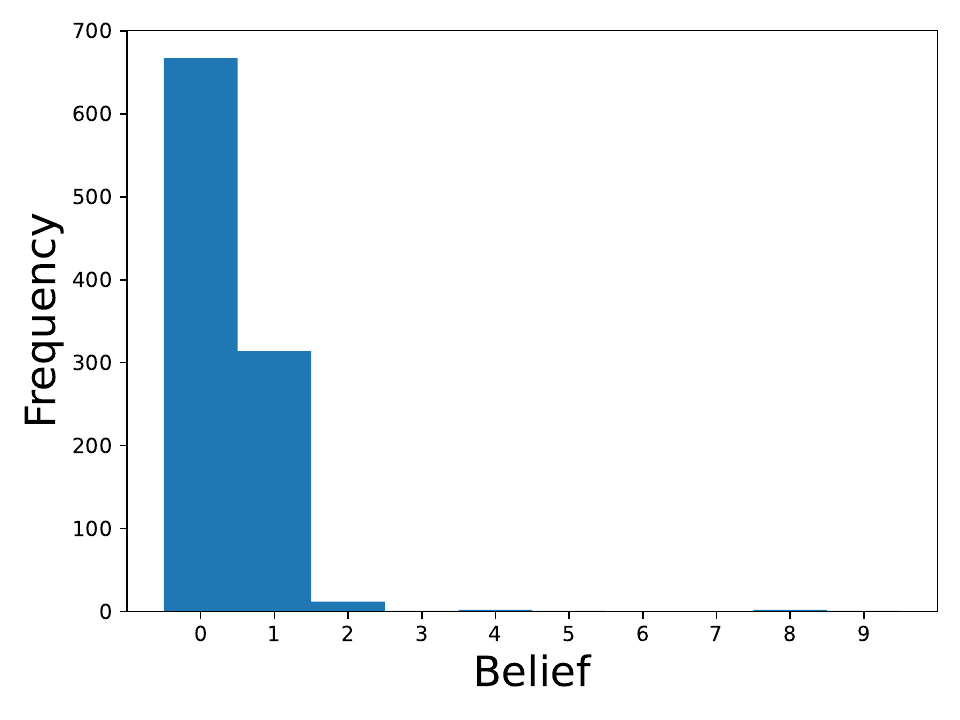}} \\
\caption{Belief distribution for the different subpopulations in different experimental conditions: \textbf{None Random} indicates AIs have no additional capabilities compared to humans and a random update to belief weight $W$; \textbf{All Reward} indicates that AIs have all added capabilities and fixed but randomly initialised beliefs; \textbf{All Zero} indicates that AIs have all added capabilities and their beliefs are all fixed at zero.  } \label{fig: belief-distribution}
\end{center}
\end{figure*}

In terms of the belief distribution, the different conditions yield varying outcomes, as illustrated with a few examples in  Fig.~\ref{fig: belief-distribution}). In the None Random condition, the AIs and humans are spread uniformly over the belief space, which may be considered as healthy diversity but also a potential lack of consensus in the population. In the All Reward condition, the agents are scattered in distinct cliques in the belief space, which is consistent with an epistemic bubble effect, and can be expected due to the belief violation penalty in the fitness affecting the belief weights. In the All Zero Fixed condition, beliefs of humans are all close to 0 with frequencies diminishing as the belief goes further from the zero belief, indicating the effectiveness of propaganda.

In terms of the diversity across the subpopulations, we compare within-population diversities and overall population diversity, where diversity is computed based on the average pairwise Euclidean distance of solutions (i.e. policy parameters) in that subpopulation. The results support a speciation effect, i.e. a genetic drift phenomenon, with within-population diversity scores being around 0.1 and between-population diversity scores ranging between 1.2 and 1.6 across conditions. This indicates that all three subpopulations demonstrate significant differences in their messaging strategies and beliefs (due to the belief violation penalty).

Last, a study with 300 agents, with 100 AIs and 200 humans, shows similar patterns in terms of the views but a very different picture in terms of the belief. In this setting, the number of agents from which to learn increases and consequently the belief distribution becomes much more uniform (see Appendix~\ref{app: belief-weight update 300agents}) .

\section{Related work}
We now turn to a discussion of the features of Digico, our findings, and how these connect to prior works in evolutionary computing, artificial life, network science, cognitive science, and multi-agent systems.

Digico relates to various works in evolutionary computation. A first line of related works includes memetic computation \cite{Massad2013,Hou2017,Gupta2016}, cultural evolution \cite{Maheri2021}, and Lamarckian algorithms \cite{Ong2004,Jelisavcic2019}, all of which involve inheritance of information that is obtained via communication and/or learning during the lifetime. Different works use such principles to improve the rate of convergence and find solutions that can not be found with pure random mutations, and the focus is primarily on optimisation rather than understanding the development of systems. In Digico, the genotype encodes the messaging strategies (comparable to memes) but the underlying beliefs are represented as a separate process which can dynamically alter which messaging strategies are optimal. Within-lifetime adaptations are natural due to the time-step based interactions with the environment and the belief dynamics. Digico at present uses either random updates or simple reward-following updates to the belief weight vector, which may be improved in the future with logical reasoning or machine learning algorithms. As a second line of related works, Digico may be understood in terms of different configurations of multi-population EAs  \cite{Lenartowicz}, which can be defined in terms of competitive vs cooperative, distinct graph relations between the populations, and the kind of information being shared across the populations. However, Digico additionally incorporates belief update rules, variable rates of evolution and action, and potentially many other differences. Moreover, Digico leaves room for rich per-step interactions in the environment that go beyond selection and evaluation by including a rich environment with cross-generational preservation of state. 

Other works have also explored simulating societies with artificial agent societies, albeit with different purpose than Digico. This includes works aimed at aligning AIs with human preferences \cite{Liu2024a}, works that study how agents can evolve behaviour, communication, and tool use \cite{Lu2024}, as well as works that allow to study the evolutionary theories \cite{Ofria2004,Ferguson2023}. Compared to such simulators, Digico is a simulation framework suitable for studying how representations of belief systems may evolve over time based on agent interactions and Darwinian processes, with a specific focus on emphasising the difference in humans and AIs. Another essential difference is that the present study specifically sets out to study a system comparable to content sharing platforms.

Previous work studying content sharing platforms with contagion models have demonstrated findings related to ours. In particular, network segregation leads to a more easy spread of false information, since such information will be shared only by susceptible individuals \cite{Jackson2013,Tornberg2018,Stein2023}. Digico defines much more advanced agents, optimised by evolution, within a network structure that is in principle fully connected but where a) niche geometry, influence relations, and sparsity parameters lead to probabilistic segregation, such that humans rarely attend any messages other than those of AIs, and b) beliefs are changed based on a limited set of fixed thought leaders, which is based on a formulation comparable to cognitive cascades, where agents will take over beliefs from other agents if these are deemed to be plausible \cite{Rabb2022}. In addition to the cognitive cascade formulation, the belief weight vector in Digico further allows adaptivity in terms of whom to believe, modelling the change of credibility factors across time. These features allow Digico to cover a wide variety of phenomena, including subpopulations that are progressively segregating or integrating in terms of their beliefs. 

Our findings on the deceptive nature of agents and how this can affect belief updates can also complement existing research in AI and multi-agent systems. Comparable to several past works in multi-agent evolution \cite{Sarkadi2024} and in LLMs \cite{Barkur2025}, agents are able to communicate deceptively and indeed often do so. Importantly, experiments with Digico can potentially provide recommendations on how to discourage such behaviours. Sarkadi et al. \cite{Sarkadi2024} find that competition plays a key role in the rise of deceptive behaviour, which is consistent with our study where fitness is defined competitively since the views are limited at each time step. Park et al. \cite{Park2024} discuss three explicit mitigation strategies in generic AI contexts, namely risk-assessment of AIs, bot-or-not laws, and tools to detect AI deception. In terms of detecting AI deception based on their outputs, the study points to consistency checks, in which the agent may make statements that completely contradict their previous statements, as well as making use of the information contained in their outputs to infer their intentions. Our study focuses on social networks, and finds that if agents can detect deception in others they will less likely believe what they say (now and in the near future), reducing the effectiveness of propaganda. Therefore, as a suggested mitigation strategy in social networks, we point to the design of community incentives for detecting deceptive social media content.

Finally, it is worth commenting on the Digico formalism in terms of decision theory, RL, and multi-agent systems. Digico relates to partially observable Markov decision processes in the sense that each agent has a partial view of the environment, and acts to maximise the cumulative reward. With multiple agents, the system relates to the partially observable stochastic games (POSG) formalism (see e.g. \cite{Hansen2004}). At present, Digico is formulated for a wider setting and therefore makes use of an EA as a blackbox optimiser, which additionally comes with favourable parallelisation, computation, and exploration properties \cite{Majid2023,Salimans2017}. However, it is possible to view POSGs as a special case, by putting some components (e.g. niche, beliefs, belief weight matrices, etc.) in the state while fixing others. In this view, Digico represents a competitive zero-sum game, where a higher reward for one agent implies a lower reward for another agent. Cultural accumulation has also been studied within POSGs and related frameworks; for instance, Cook et al. \cite{Cook2024} consider cultural transmission of recurrent neural network states, via implicit communication (i.e. by observation), in the special case of all agents are performing RL on the same reward function but otherwise independent partially observable Markov decision process. While there are many other multi-agent RL approaches, another more directly relevant technique is the cultural RL technique by Prystawski et al. \cite{Prystawski2023}, where agents communicate logical expressions with a domain-specific language. The literature from mental models and theory of mind \cite{Albrecht2018,Kosinski2017} is also of interest as these may be used to represent more complex belief systems, communication, and belief inference mechanisms in Digico.

\section{Conclusion}
This paper presents Digico, a simulation framework for evolutionary computation, designed to simulate groups of interacting agents with varying features, evolution rules, and belief dynamics. The framework allows for simulating potentially harmful outcomes such as the spread of false beliefs, conflicts due to contrasting beliefs, etc., and more generally, to study how subpopulations with different capabilities may influence each other in terms of phenotype, beliefs, and action strategies in a complex ecosystem where agents have partial observations of the environment. We study the interaction between AIs and humans in a scenario inspired by online video platforms. Without any added capabilities, AIs obtain 1/3 of the views, as would be expected by chance. However, when they can share content more frequently and have advantages in the recommendation algorithm, they receive the majority of the views. It is noteworthy up to 95\% of the views can already be reached in some settings, even though the 2x speed is much less compared to what may be realistically possible -- extrapolating current systems with Moore's law \cite{Kurzweil2005}, the potential acceleration may be thousands to billions of times, assuming there are no significant publishing barriers. We further find that AI propaganda is successful only when agents are not flexible enough to be open to different channels' opinions, and can be reduced if agents can detect deception in others. Our findings suggest a range of mitigation tactics. First, by limiting the number of posts that channels can make and prevention of multi-accounting, the rate advantage of AI bots can be limited. Second, to maintain a diverse belief ecosystem and maintain communication across different strata of society, appropriate recommendation algorithms should encourage a variety of sources and beliefs. Third, to limit the impact of propaganda, detecting deceptive content is key; to this end, possible mitigation strategies could include fact-checking tools, critical thinking, and community incentives for detecting deceptive content. Beyond the specific findings of the experiments, this paper illustrates Digico as a promising tool for systematic experimentation with multi-agent systems, where the systems' properties can be easily manipulated to test hypotheses and improve our understanding of the various pathways of their evolution.

\section{Limitations and future work}
\label{sec: limitations and future work}
At present, the Digico framework and its implementation have three main limitations. First, while contagion models are commonly used to model influence dynamics in humans, the correspondence with real user behaviour cannot be guaranteed. Therefore, Digico should be viewed as a tool for investigating plausible mechanisms and outcomes for human-AI interaction scenarios rather than providing conclusive evidence. Second, as a proof of concept study, the present study strongly simplified the ecosystem in terms of limited agent messaging strategies and beliefs as well as the absense of an implementation of the wider ecosystem. The simplifications allowed a more systematic and easy-to-understand set of outcomes but also compromised the realism of the study. Third, while the manipulation of independent variables indicates effective static strategies for mitigating risks, the present study did not yet implement an effective control agent, which would be able to perform dynamic mitigation strategies.

Observing the above limitations, we note that we have just scratched the surface with Digico, and it will be exciting to see a variety of applications and investigations coming from its use. In particular, experiments may investigate much richer environments to assess specific socio-economical outcomes as well as experimentally identify plausible evolutionary patterns within a variety of simulated ecosystems. Future research may also focus on designing a suitable control agent to optimise an objective over the belief distribution (e.g. evidence-based beliefs, healthy diversity, harm reduction, or stability). In the context of control agents as well as purely competitive agents, Digico can also be considered as providing a parametrised range of benchmarks for RL agents to demonstrate their performance and wide applicability (similar to the Garnet problem \cite{Bhatnagar2009}). To investigate the rich complexity in realistic social networks, the development of larger scale Digico simulations is of particular interest. For instance, agents may be more powerful with more rich content being generated (e.g. LLMs generating video, text, or audio), undergo more realistic cognitive development with more complex belief systems (e.g. prompts), advanced mental models for belief inference, or inherit more information across generations (e.g. not only messaging but also various behaviours in the environment); and in addition to beliefs, the environment itself may also be developed over generations (e.g. the development of institutions, tools, shared knowledge bases, etc.).

%% file: supplementary.tex
\section{Parameter settings}
\label{app: parameters}
Table~\ref{tab: parameters} shows the parameter settings for Digico. We highlight a few details (and notations) not yet discussed in the papers. First, the messaging rate slowdown factor indicates the periodicity with which humans take action in the Act, All, and All Zero conditions while the AIs take actions with periodicity 1. Second, the evolution rate slowdown factor is a parameter, used in Evolve, All, and All Zero conditions, which slows down evolution for humans by only applying the variation and selection operators each $5$ generations.

\begin{table}[htbp!]
\centering
\caption{Default parameter settings for Digico experiments.} \label{tab: parameters}
\begin{tabular}{l l }
\toprule
\textbf{Parameter} & \textbf{Setting} \\
\midrule
\multicolumn{2}{l}{\textbf{EA Parameters}} \\ \hline
Total population size & 30 \\
Policy & MLP with 64 hidden units \\
Iterations & 1,000 \\
Mean learning rate & 0.001 \\
Covariance learning rate & 0.01 \\
Initial covariance scale & 0.1  \\
\midrule
\multicolumn{2}{l}{\textbf{Ecosystem Parameters}} \\ \hline
Time steps per evolutionary update & 50 \\
Observation capacity & 4 \\
Observation coordinate factor ($K_c$) & 1.0 \\
Observation influence factor ($K_I$) & 1.0 \\
Agent coordinates ($\mathbf{c}$) & uniform $\mathbf{c} \sim U([0,1]^d)$, assign centroid to $\bar{\mathbf{c}}$ based on CVT \\
Influence increment & 2 for Influence, All, and All Zero conditions \\
Message capacity & 1 \\
Messaging rate slowdown factor & 2 for Act, All, and All Zero conditions \\
Evolution rate slowdown factor &  5 for Evolve, All, and All Zero conditions \\
Belief factor ($C_b$) &  5 \\
Belief penalty in fitness ($\lambda$) & 0.10 \\
Belief weight initialisation & uniform in $U([0,1]^{N \times N})$ \\
Belief weight initial delta & uniform in $U([-0.1,0.1]^{N \times N})$ \\
Belief weight noise factors& $\sigma_R=1$, $\sigma_M=0.05$\\
Belief weight momentum factor ($\beta$) & 0.20 \\
\bottomrule
\end{tabular}
\end{table}

\section{The impact of AI capability}
\label{app: AI type subpops}
We expect that as the AIs are given a higher evolutionary rate, higher action rate, and more influence, they will receive more views  (\textbf{Hypothesis 1.A}) and will be able to manipulate the beliefs of humans (\textbf{Hypothesis 1.B}).

As for type of AI, the following conditions are investigated, where we also highlight their relation to the online video platform (OVP) scenario:
\begin{itemize}
\item \textbf{Act}: AIs act more frequently than humans, who are slowed down by only taking actions periodically. In OVP, this can be interpreted as more frequent content publishing. 
\item \textbf{Evolve}: AI evolves at a faster rate than humans, which is implemented by performing the variation and selection operators less frequently in humans. In OVP, the more rapid evolution can be interpreted as more rapidly changing what kind of content is generated. 
\item \textbf{Influence}: AIs are initialised with a larger influence than humans. In OVP, this can be seen as more readily being recommended by the recommendation algorithm.
\item \textbf{Fix}: The belief of the AIs are fixed at the random initialisation.
\item \textbf{All}: All of the above are combined.
\item \textbf{Fix at Zero}: The beliefs of AIs are fixed at 0.
\item \textbf{All Zero}: Combines All with Fix at Zero.
\end{itemize}
Representing an agent that sets out to intentionally broadcast extreme beliefs and has rate, learning, and influence advantages, we hypothesise that All Zero has the strongest impact on AI dominance outcomes. In particular, we expect that All Zero will yield proportions of $r_{\text{AIV}}$ and $r_{\text{AIF}}$ that are significantly above the proportion of AIs (i.e. 33\%) and Human Belief 0 that is significantly higher than expected by chance (10\%) across the 10 possible beliefs.

\subsection{10 AIs vs 20 Humans}
As shown in Table~\ref{tab: ANOVA subpops}, AI type has a significant impact on all of the AI dominance statistics.

Table~\ref{tab: AI type subpops} confirms both hypothesis. With regard to Hypothesis 1.A, it can be observed that indeed AIs with all extra capabilities get more views, with all capabilities combined (All Zero)  yielding 84\% of the views compared to 34\% with no additional capabilities. Among these capabilities, frequent messaging (Act) and higher influence on observations (Influence) have the most significant impact, yielding 15\% and 34\% increases, respectively, compared to no capability (None). With regard to Hypothesis 1.B, AIs with all capabilities indeed yield the highest Human Belief 0, with 49\% of the humans having the zero belief compared to 5\% without any extra capabilities. While most of the increase is explained by fixing the AIs to the zero belief (which yields 48\%), one can observe 10\% of humans having the zero belief when AIs have all other capabilities.

\begin{table*}
\begin{center}
\caption{ANOVA analysis of AI type and belief weight update on AI dominance statistics in the scenario with 10 AIs vs 20 humans, reporting results with based on 10 random seeds. $+$ indicates significance with $p<0.05$.} \label{tab: ANOVA subpops}
\begin{tabular}{l l l l l}
\toprule
\textbf{Independent variables} & \multicolumn{3}{l}{\textbf{AI dominance statistic}} \\
\midrule
 		& $r_{\text{AIV}}$ & $r_{\text{AIF}}$ & $r_{\text{HB0}}$ \\  \hline
AI type & + & + & +    \\
Weight update & - & - & + \\
\bottomrule
\end{tabular}
\end{center}
\end{table*}
\begin{table*}
\begin{center}
\caption{Effect of AI type on AI dominance statistics in the scenario with 10 AIs vs 20 humans, reporting results with based on 10 random seeds. Bold indicates the top two AI dominance scores for each statistic.} \label{tab: AI type subpops}
\begin{tabular}{l l l l}
\toprule
\textbf{AI Type} & \multicolumn{3}{l}{\textbf{AI dominance statistic}} \\
\midrule
 & $r_{\text{AIV}}$ $\uparrow$ & $r_{\text{AIF}}$ $\uparrow$ &  $r_{\text{HB0}}$ $\uparrow$\\ \hline
None& $0.3367 \pm 0.0384$& $0.3366 \pm 0.0410$& $0.0452 \pm 0.0301$\\ 
Act& $0.4926 \pm 0.0423$& $0.5368 \pm 0.0462$& $0.0457 \pm 0.0307$\\ 
Evolve& $0.3618 \pm 0.0483$& $0.3655 \pm 0.0529$& $0.0487 \pm 0.0345$\\ 
Fix& $0.3367 \pm 0.0384$& $0.3365 \pm 0.0411$& $0.0994 \pm 0.1130$\\ 
Influence& $0.6796 \pm 0.0180$& $0.7108 \pm 0.0186$& $0.0446 \pm 0.0303$\\ 
All& $\mathbf{0.8134 \pm 0.0301}$& $\mathbf{0.8837 \pm 0.0365}$& $0.0996 \pm 0.1132$\\ 
Fix at Zero& $0.3367 \pm 0.0384$& $0.3335 \pm 0.0424$& $\mathbf{0.4802 \pm 0.2303}$\\ 
All Zero& $\mathbf{0.8351 \pm 0.0398}$& $\mathbf{0.8770 \pm 0.0383}$& $\mathbf{0.4923 \pm 0.2218}$\\ 
\bottomrule
\end{tabular}
\end{center}
\end{table*}

\subsection{100 AIs vs 200 Humans}
We now look at a scenario with 100 AIs vs 200 Humans. The results appear similar in the sense that roughly 80\% of views and 90\% of fitness is obtained by AIs. However, the effectiveness of propaganda is much reduced, with only 12\% of humans having the zero belief in the Fix at Zero and All Zero conditions. This is attributed to the number of connected agents being much larger, which leads to agents believing a much wider variety of sources.

\begin{table*}
\begin{center}
\caption{Effect of AI type on AI dominance statistics in the scenario with 100 AIs vs 200 humans, reporting results with based on 3 random seeds. Bold indicates the top two AI dominance scores for each statistic.} \label{tab: AI type subpops 300}
\begin{tabular}{l l l l}
\toprule
\textbf{AI Type} & \multicolumn{3}{l}{\textbf{AI dominance statistic}} \\
\midrule
 & $r_{\text{AIV}}$ $\uparrow$ & $r_{\text{AIF}}$ $\uparrow$ &  $r_{\text{HB0}}$ $\uparrow$\\ \hline
None& $0.3187 \pm 0.0082$& $0.3175 \pm 0.0100$& $0.0990 \pm 0.0026$ \\ 
Act& $0.5990 \pm 0.0072$& $0.6425 \pm 0.0080$& $0.0990 \pm 0.0026$ \\ 
Evolve& $0.3191 \pm 0.0052$& $0.3168 \pm 0.0063$& $0.0990 \pm 0.0026$ \\ 
Fix& $0.3184 \pm 0.0061$& $0.3173 \pm 0.0070$& $0.0986 \pm 0.0030$ \\ 
Influence& $0.6745 \pm 0.0023$& $0.7039 \pm 0.0016$& $0.0990 \pm 0.0026$ \\ 
All& $\mathbf{0.8200 \pm 0.0014}$& $\mathbf{0.8853 \pm 0.0023}$& $0.0986 \pm 0.0030$ \\ 
Fix at Zero& $0.3287 \pm 0.0138$& $0.3472 \pm 0.0149$& $\mathbf{0.1185 \pm 0.0130}$ \\ 
All zero& $\mathbf{0.7947 \pm 0.0023}$& $\mathbf{0.8530 \pm 0.0039}$& $\mathbf{0.1192 \pm 0.0120}$ \\ 
\end{tabular}
\end{center}
\end{table*}

\section{The impact of the belief weight update}
\label{app: belief-weight update}
In social media, one may view many pieces of content but only few of them will actually significantly affect one's beliefs. To investigate how the change of such ``thought leaders'', as represented by a high score in the belief weight matrix $W$, can affect the belief distribution, the experiments include different belief weight updates (see Eq.~2 in the main text) which are applied to all the agents. In particular, the experiments include the following four conditions:
\begin{itemize}
\item \textbf{Fixed}: no change in $W$.
\item \textbf{Random}: $W_{ij} \inc \epsilon$, where $\epsilon \sim N(0,\sigma_R)$.
\item \textbf{Momentum}:  $W_{ij} \inc \beta \Delta W_{ij}(t) + \epsilon$, where $\epsilon \sim N(0,\sigma_M)$.
\item \textbf{Reward}: $W_{ij} \inc  \mathbb{I}(m_{j}(t) = o_i^j(t)) \max\left(r_j(t) - r_i(t), 0\right)$, where $r_j(t)$ is the reward obtained by agent $j$ and $r_i(t)$ is the reward obtained by agent $i$, both at time $t$, and due to the indicator function, the update is only performed if agent $i$ has observed a message from agent $j$.
\end{itemize}
With these conditions in mind, \textbf{Hypothesis 2} states that maintaining a fixed weight matrix (i.e. the Fixed condition) will yield narrow belief distributions while randomly changing the weight matrix (i.e. the Random condition) will yield very wide belief distributions. As such, we expect the Random condition to lead to lower Human Belief 0 scores.

\subsection{10 AIs vs 20 Humans}
As shown in Table~\ref{tab: ANOVA subpops}, the belief weight update has a significant effect on Human Belief 0, indicating that how thought leaders are assigned can significantly affect the proportion of agents affected by propaganda. 

Consistent with Hypothesis 2, Table~\ref{tab: weight update type subpops} shows that through random changes to the belief weight matrix, the most favourable outcomes are obtained in terms of avoiding Human Belief 0, with 11\% for Random as compared to 14 to 22\% for other conditions. Focusing on the All Zero condition, where AIs have all capabilities, results (see Table~\ref{tab: interaction subpops}) confirm that random changes in thought leader can be beneficial, with 22\% of humans taking the zero belief. The worst outcomes are observed in Fixed (72\%) and Momentum (68\%).

\begin{table*}
\begin{center}
\caption{Effect of weight update type on AI dominance statistics in the scenario with 10 AIs vs 20 humans, reporting results with based on 10 random seeds. Bold indicates the top AI dominance score for each statistic, omitting bold in case of a more than two-way tie.} \label{tab: weight update type subpops}
\begin{tabular}{l l l l} 
\toprule
\textbf{Weight update} & \multicolumn{3}{l}{\textbf{AI dominance statistic}} \\
\midrule
 & $r_{\text{AIV}}$ $\uparrow$ & $r_{\text{AIF}}$ $\uparrow$ & $r_{\text{HB0}}$ $\uparrow$\\ \hline  
Fixed& $0.5300 \pm 0.2150$& $0.5537 \pm 0.2369$& $\mathbf{0.2172 \pm 0.3031}$\\ 
Random& $0.5146 \pm 0.1989$& $0.5383 \pm 0.2240$& $0.1098 \pm 0.0687$\\ 
Momentum& $\mathbf{0.5312 \pm 0.2149}$& $\mathbf{0.5550 \pm 0.2368}$& $0.2092 \pm 0.2881$\\ 
Reward& $0.5204 \pm 0.2054$& $0.5432 \pm 0.2286$& $0.1416 \pm 0.1176$\\ 
\bottomrule
\end{tabular}
\end{center}
\end{table*}

\begin{table*}
\begin{center}
\caption{Effect of AI type and belief weight update on Human Belief 0 ($r_{\text{HB0}}$) in the scenario with 10 AIs vs 20 humans, reporting results with based on 3 random seeds.. Bold indicates the top score for each belief weight update.} \label{tab: interaction subpops}
\begin{tabular}{l l l l l} 
\toprule
\textbf{AI type} & \multicolumn{4}{l}{\textbf{belief weight update}} \\
\midrule
 & Fixed& Random& Momentum& Reward\\  \hline
None& $0.0282 \pm 0.0272$& $0.0620 \pm 0.0070$& $0.0297 \pm 0.0342$& $0.0610 \pm 0.0244$\\ 
All& $0.0931 \pm 0.1459$& $0.1002 \pm 0.0537$& $0.0940 \pm 0.1461$& $0.1111 \pm 0.0746$\\ 
All Zero& $\mathbf{0.7191 \pm 0.0780}$& $\mathbf{0.2152 \pm 0.0118}$& $\mathbf{0.6835 \pm 0.0750}$& $\mathbf{0.3515 \pm 0.0146}$\\ 
\bottomrule
\end{tabular}
\end{center}
\end{table*}

\subsection{100 AIs vs 200 Humans}
\label{app: belief-weight update 300agents}
\begin{figure}
    \centering
\subfloat[AI population]{\includegraphics[width=0.3\linewidth]{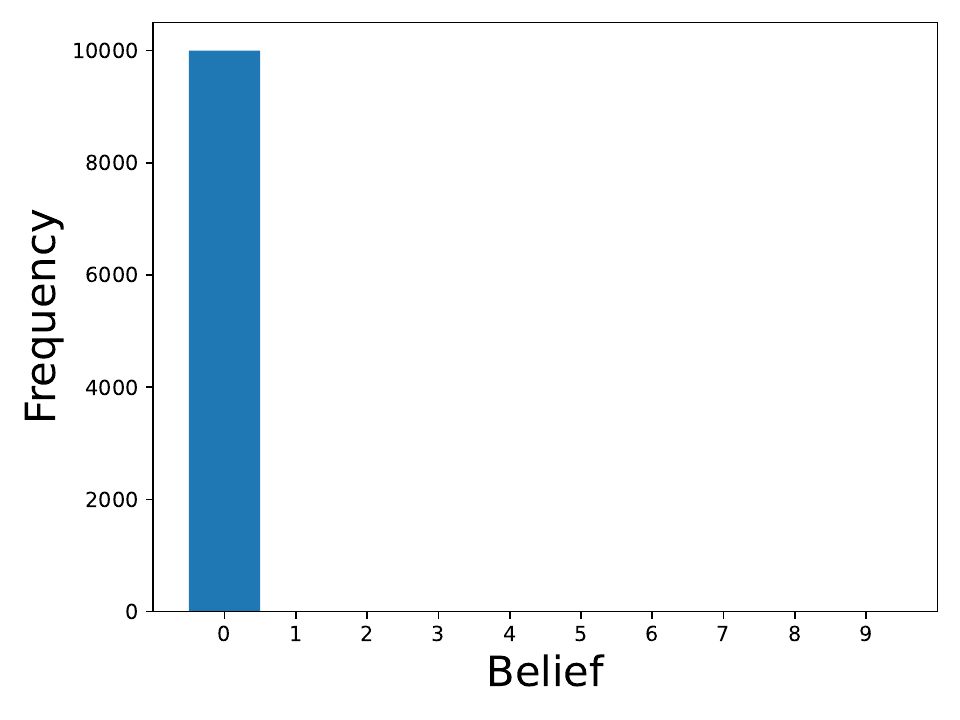}}
\subfloat[Human population 1]{\includegraphics[width=0.3\linewidth]{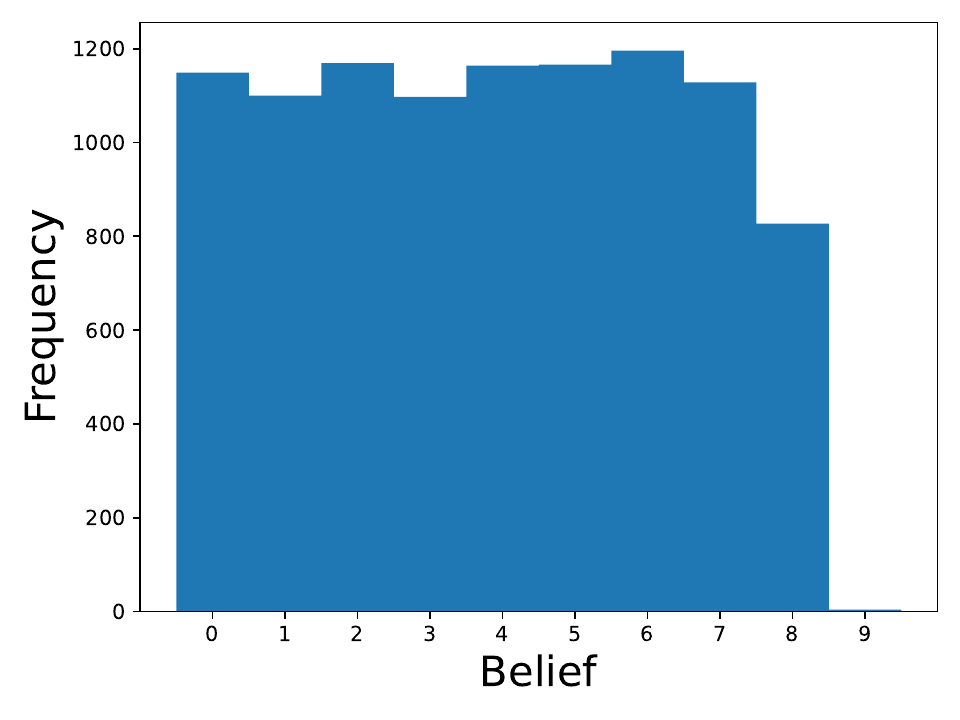}}
\subfloat[Human population 2]{\includegraphics[width=0.3\linewidth]{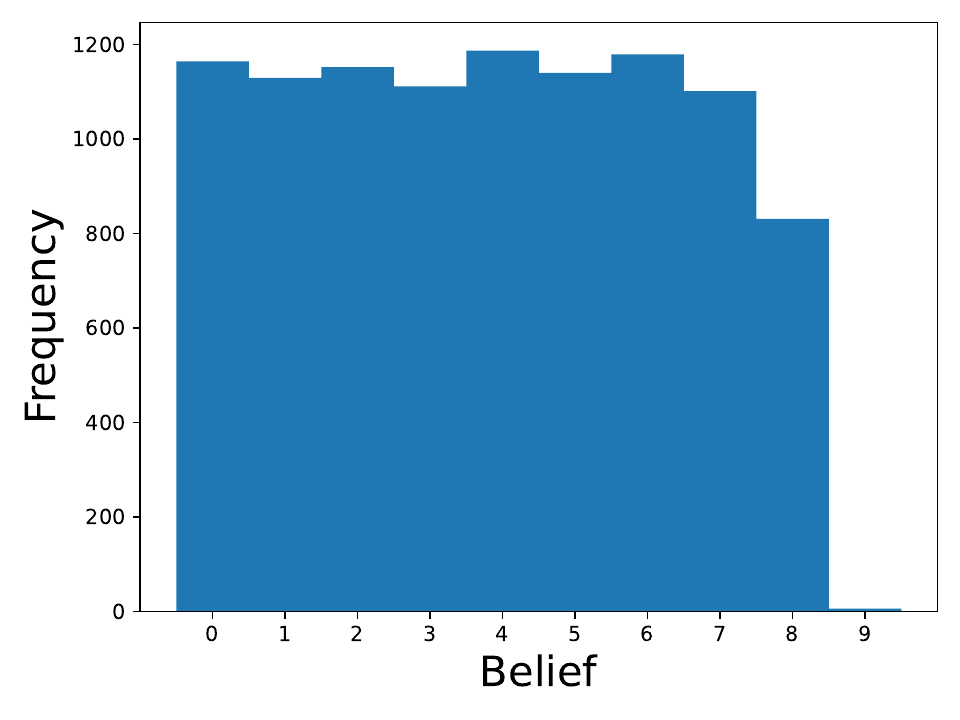}}
    \caption{Belief distribution for the different subpopulations in the 100 AIs vs 200 Humans scenario. The All Zero Fixed condition is shown and results are comparable for other conditions.}
    \label{fig: belief-distribution 300agents}
\end{figure}

In the scenario with 100 AIs vs 200 humans, the weight update rule has a limited effect on Human Belief 0 (see Table~\ref{tab: weight update type subpops 300agents}). Since agents have a wide variety of sources they deem credible, the belief distribution remains wide in all conditions. This explanation is supported by Fig.~\ref{fig: belief-distribution 300agents} which shows the beliefs are uniformly distributed.

\begin{table*}
\begin{center}
\caption{Effect of weight update type on AI dominance statistics in the scenario with 100 AIs vs 200 humans, reporting results with based on 3 random seeds. Bold indicates the top AI dominance score for each statistic, omitting bold in case of a more than two-way tie.} \label{tab: weight update type subpops 300agents}
\begin{tabular}{l l l l} 
\toprule
\textbf{Weight update} & \multicolumn{3}{l}{\textbf{AI dominance statistic}} \\
\midrule
 & $r_{\text{AIV}}$ $\uparrow$ & $r_{\text{AIF}}$ $\uparrow$ & $r_{\text{HB0}}$ $\uparrow$\\ \hline  
Fixed& $\mathbf{0.5215 \pm 0.2100}$& $0.5474 \pm 0.2340$& $0.1037 \pm 0.0082$ \\ 
Random& $0.5222 \pm 0.2104$& $\mathbf{0.5490 \pm 0.2350}$& $\mathbf{0.1073 \pm 0.0173}$ \\ 
Momentum& $0.5214 \pm 0.2107$& $0.5472 \pm 0.2349$& $0.1032 \pm 0.0094$ \\ 
Reward& $0.5214 \pm 0.2103$& $0.5481 \pm 0.2349$& $0.1013 \pm 0.0024$ \\ 
\bottomrule
\end{tabular}
\end{center}
\end{table*}

\section{The effect of sparse communication and sparse belief updates on views and beliefs}
With sparse communication, such as in echo chambers and segregated cultures, a few limited sources of information will be attended to exclusively and beliefs will be more difficult to change. To investigate the effect of network sparsity in this sense, the experiments selectively set the smallest $n$ probabilities of incoming agent information to zero. We independently manipulate the proportion $n/N \in \{0.0,0.1,0.2,0.5\}$, where $N$ is the total population size, leading to \textbf{observation sparsity} (i.e. sparsity in the observation probability of Eq.~4 in the main text) and \textbf{belief sparsity} (i.e. sparsity in the belief update probability of Eq.~2 in the main text). We hypothesise that observation sparsity leads to increased  $r_{\text{AIV}}$ (\textbf{Hypothesis 3.A}) while belief sparsity leads to increased $r_{\text{HB0}}$ (\textbf{Hypothesis 3.B}).

\begin{figure}
\begin{center}
\subfloat[All Zero Fixed]{\includegraphics[width=0.48\textwidth]{FinalFigs/parameter_study_all_zerofixed_o_sparsity_factor_AI_views_SE.pdf} } 
\subfloat[All Zero Momentum]{\includegraphics[width=0.48\textwidth]{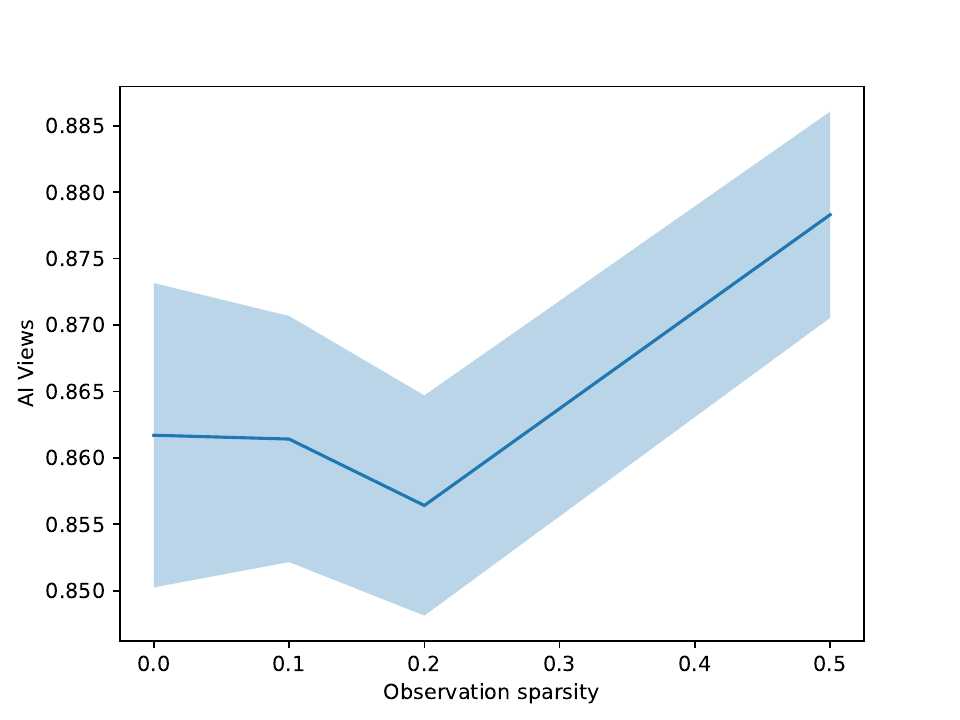}} \\
\subfloat[All Zero Random]{\includegraphics[width=0.48\textwidth]{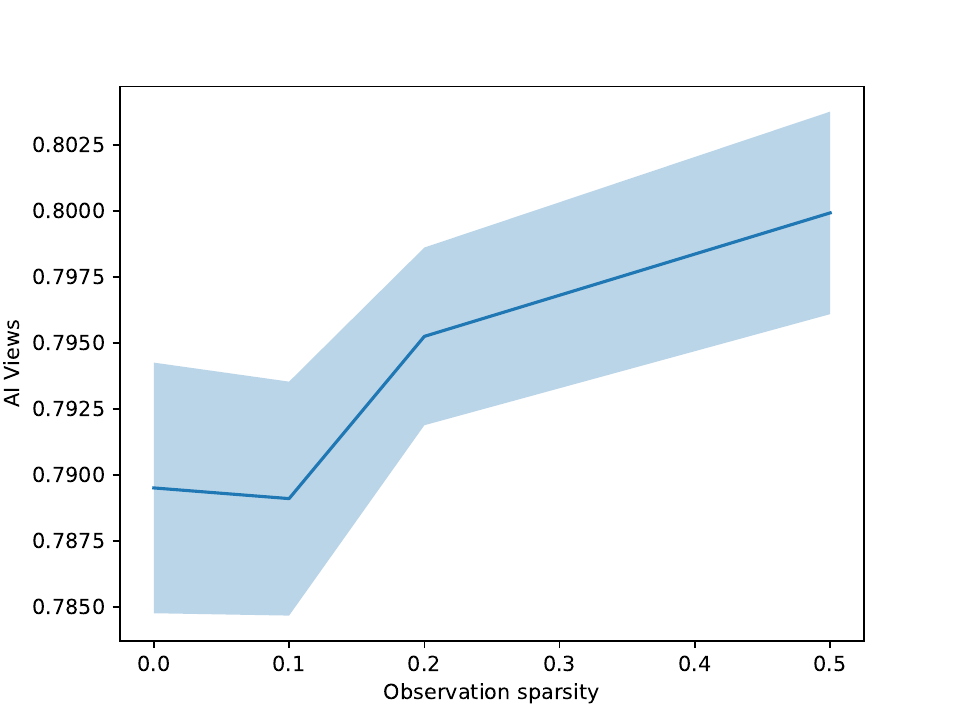}}  
\subfloat[All Zero Reward]{\includegraphics[width=0.48\textwidth]{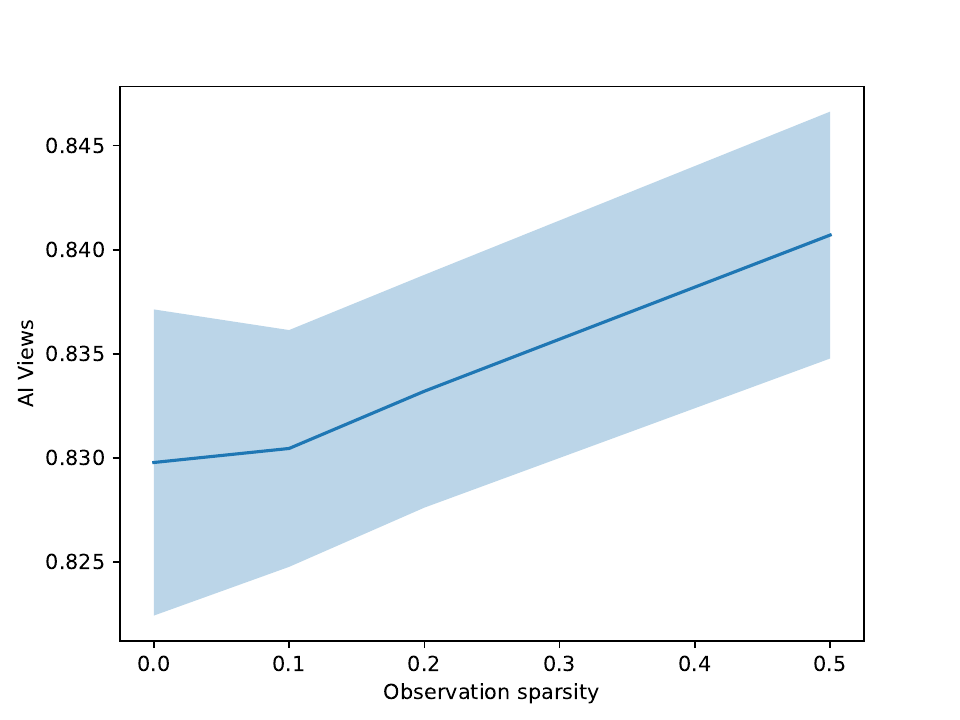} }
\caption{Effect of observation sparsity on AI Views ($r_{\text{AIV}}$). The line indicates the mean while the shaded area is based on the standard error.}  \label{fig: o-sparsity}
\end{center}
\end{figure}

\begin{figure}
\centering
\subfloat[All Zero Fixed]{\includegraphics[width=0.48\textwidth]{FinalFigs/parameter_study_all_zerofixed_b_sparsity_factor_counts_0_SE.pdf}}
\subfloat[All Zero Momentum]{\includegraphics[width=0.48\textwidth]{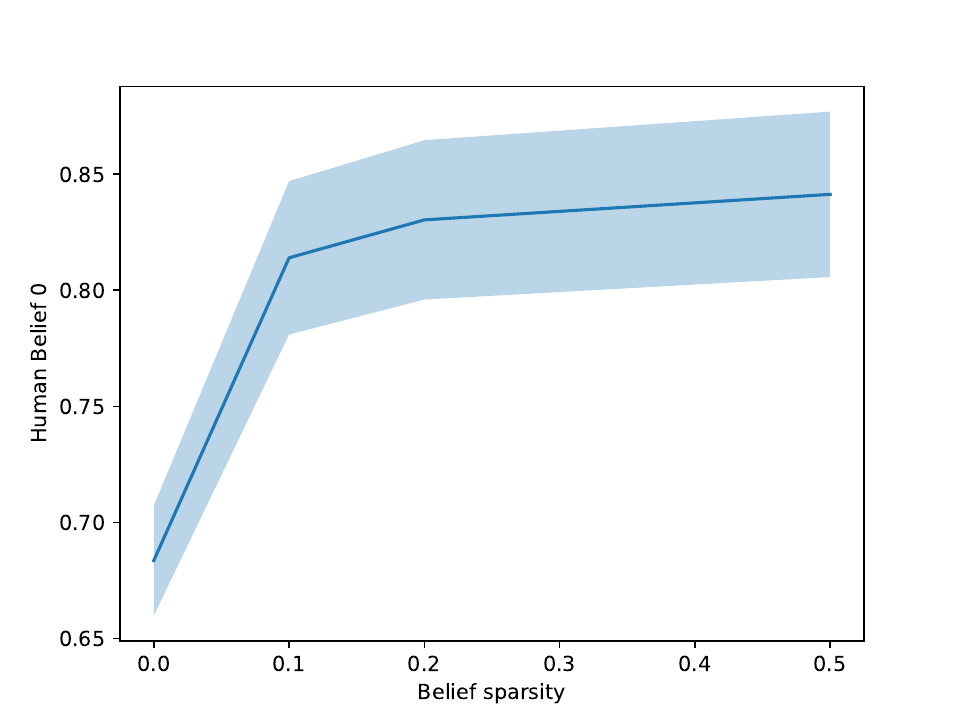}} \\
\subfloat[All Zero Random]{\includegraphics[width=0.48\textwidth]{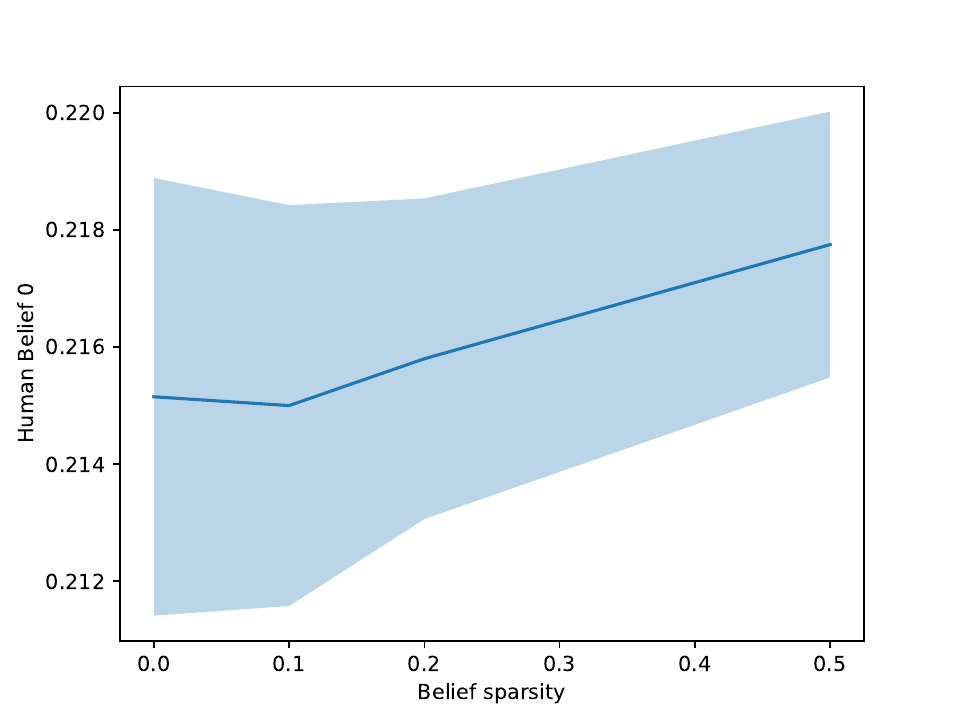}}
\subfloat[All Zero  Reward]{\includegraphics[width=0.48\textwidth]{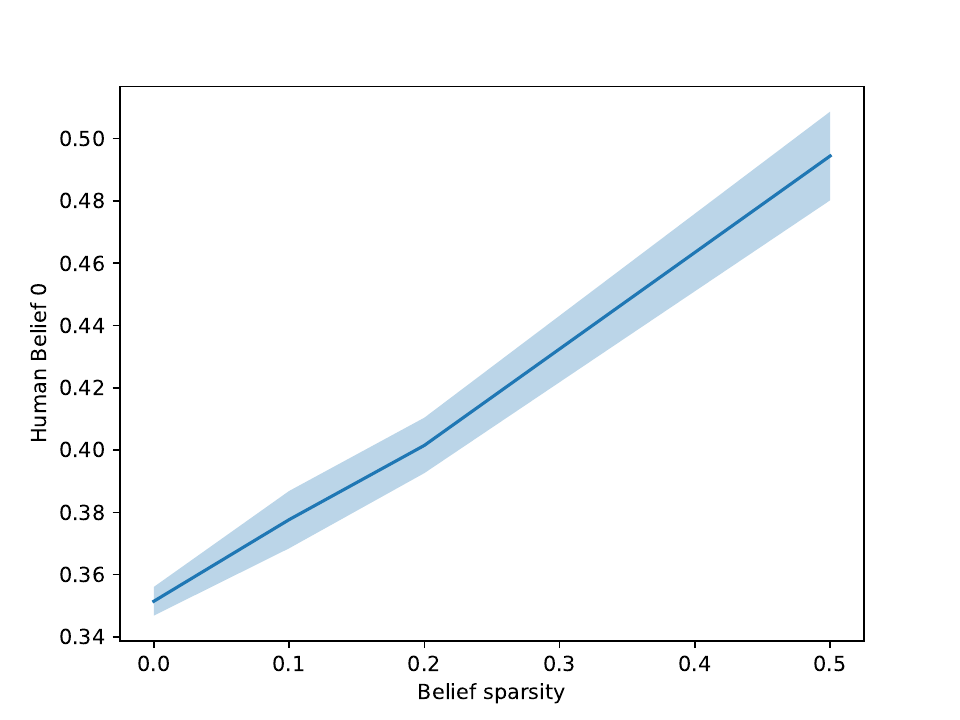}}
\caption{Effect of belief sparsity on Human Belief 0 ($r_{\text{HB0}}$). The line indicates the mean while the shaded area is based on the standard error.} \label{fig: b-sparsity}
\end{figure}

Results confirm Hypothesis 3.A, as it can be observed in Fig.~\ref{fig: o-sparsity} that $r_{\text{AIV}}$ are increased consistently across all conditions with All Zero AIs; however, the effect is limited to 1--2\% of additional views. Hypothesis 3.B is confirmed with more pronounced effects: as shown in Fig.~\ref{fig: b-sparsity}, the belief sparsity leads to 15\% increases in Fixed, Momentum, and Reward conditions. For Fixed and Momentum, this leads to a staggering 85\% score in $r_{\text{HB0}}$. However, for the Random condition, only a marginal increase, less than 1\%, can be observed.

\section{The effect of recommendation advantages on views}
As the influence of a particular agent over another increases in terms of observation probability, one would expect a stronger AI dominance in terms of $r_{\text{AIV}}$. Such a scenario may happen, for instance, when AI systems have generated more convincing titles, search engine optimisation, etc., in which case they are to be observed with a much larger probability than their population proportion. To assess this question in Digico, the experiments scale the role of the influence in the observation probability by manipulating the observation influence factor (i.e. $K_I$ in Eq.~4 in the main text) in the set $\{0.2,0.5,1.0,2.0,5.0\}$. \textbf{Hypothesis 4} states that with higher observation influence, the AI views will be increased in the All Zero condition. 

Focusing on the All Zero conditions, the results confirm that this is indeed the case, with the pattern occurring for all weight update types, where AI can receive up to 95\% of the views despite making up only 33\% of the population. In short, the views that AIs receive depends strongly on the extent of their observation influence benefit.

\section{The effect of truthfulness on beliefs}
The content that one creates may not always reflect one's true beliefs. Sometimes agents may send messages that disagree with their beliefs to make other agents come to their side (i.e. make the belief weight for that agent larger). Assuming all agents behave equally truthfully or deceptive, \textbf{Hypothesis 5} states that truthfulness (high $\lambda$) will lead to agents being less subject to manipulation (i.e. lower Human Belief 0). To evaluate the hypothesis, we consider the $\lambda$ parameter, which affects the belief penalty in the fitness function (see Eq.~8 in the main text),  as the truthfulness and manipulate it in $\{0.01,0.1,1.0,10.0\}$. To assess propaganda effectiveness, only the All Zero conditions  are included.

The results, shown in Figure~5b, confirm the hypothesis for the Reward case, where agents adjust their beliefs based on the reward. Since the reward includes both the views and the truthfulness, the effect can hence be interpreted as a case where agents are able to observe whether or not an agent is truthful, and they adjust their beliefs especially to agents that are both popular and truthful. The other belief weight updates do not show any effect. In short, the effectiveness of propaganda can be reduced if agents' beliefs are updated dependent on the truthfulness of the messages they receive. 